\documentclass[11pt]{article}

\usepackage[preprint]{acl}

\usepackage{times}
\usepackage{latexsym}

\usepackage[T1]{fontenc}
\usepackage[utf8]{inputenc}

\usepackage{microtype}

\usepackage{inconsolata}
\usepackage{amsmath}
\usepackage{amssymb}
\usepackage{booktabs}       % professional-quality tables
\usepackage{arydshln}
\usepackage{graphicx}
\usepackage[table]{xcolor}
\usepackage[most]{tcolorbox}
\usepackage{multirow}
\usepackage{multicol}
\usepackage{bbding}
\usepackage{soul}
\usepackage{xcolor}
\sethlcolor{green!20}
\newcommand{\sys}{\textsc{TRUST}}
\newtcolorbox[auto counter]{promptbox}[2][]{%
  breakable,
  enhanced,
  colback=gray!5!white,
  colframe=gray!75!black,
  boxrule=0.5mm,
  width=\textwidth,
  arc=3mm,
  auto outer arc=true,
  fonttitle=\bfseries,
  title=Box~\thetcbcounter: #2,
  #1
}

\title{Exploring Agentic Tool-Calling Decisions via Uncertainty-Aligned Reinforcement Learning}

\author{
 \textbf{Yijin Zhou\textsuperscript{1,2,3}},
 \textbf{Linqian Zeng\textsuperscript{1}}, 
 \textbf{Xiaoya Lu\textsuperscript{1,2}},
 \textbf{Wenyuan Xie\textsuperscript{1}}, \\
 \textbf{Dongrui Liu\textsuperscript{2,$\dagger$}},
 \textbf{Junchi Yan \textsuperscript{1,3}},
 \textbf{Jing Shao\textsuperscript{2,$\dagger$}}
\\
 \textsuperscript{1}Shanghai Jiao Tong University, China \\
 \textsuperscript{2}Shanghai Artificial Intelligence Laboratory, China \\
 \textsuperscript{3}Shanghai Innovation Institute, China
}

\begin{document}

\maketitle

\begin{abstract}
Large language model (LLM)-based agents often make suboptimal tool-use decisions, including unsupported tool invocation and hallucinated direct responses, which may accumulate errors throughout multi-step interactions. Existing approaches mainly improve these behaviors through inference-time correction or coarse-grained reward signals based on decision outcomes and structured checklists, leaving the uncertainty characteristics of agent decisions underexplored. We observe that decision-oriented reinforcement learning tends to weaken the uncertainty separation between correct and incorrect actions, resulting in overconfident mistakes and weaker exploration signals. Therefore, we propose \sys, which incorporates uncertainty quantification into reward design as a repulsive force for maintaining uncertainty separation, and labels lightweight key-turn annotations for unified post-training of multi-turn trajectories. Experimental results across diverse tool-use benchmarks show that \sys~consistently enhances both decision quality and agent performance while maintaining more reliable uncertainty estimates during optimization.

\textbf{Code} -- \href{https://github.com/yjzscode/TRUST}{https://github.com/yjzscode/TRUST}
\end{abstract}
\section{Introduction}

Large language model (LLM)-based agents extend the capabilities of language models by invoking external tools for knowledge retrieval, computation, and interaction with real-world environments \citep{yao2023react,schick2023toolformer,qin2024toolllm, lu2026bench, chen2026geometrically}. However, agents frequently exhibit tool-calling decision failures at specific action turns, either invoking tools when tool use is unsupported or unnecessary, or fabricating direct answers without issuing the required tool call \citep{ross2025when2call, healy2026toolselection, zhou2026infa}. Unlike conventional textual hallucinations in LLMs, tool-calling decision failures can corrupt intermediate states and propagate errors across subsequent interaction turns, thereby incurring financial costs, execution failures, and information leakage in real-world agentic tasks \citep{lin2025agenthallusurvey, su2025autonomyrisk, zhang2025agentsafetybench, lu2026homeguard}.
\begin{figure}
    \centering
    \includegraphics[width=\linewidth]{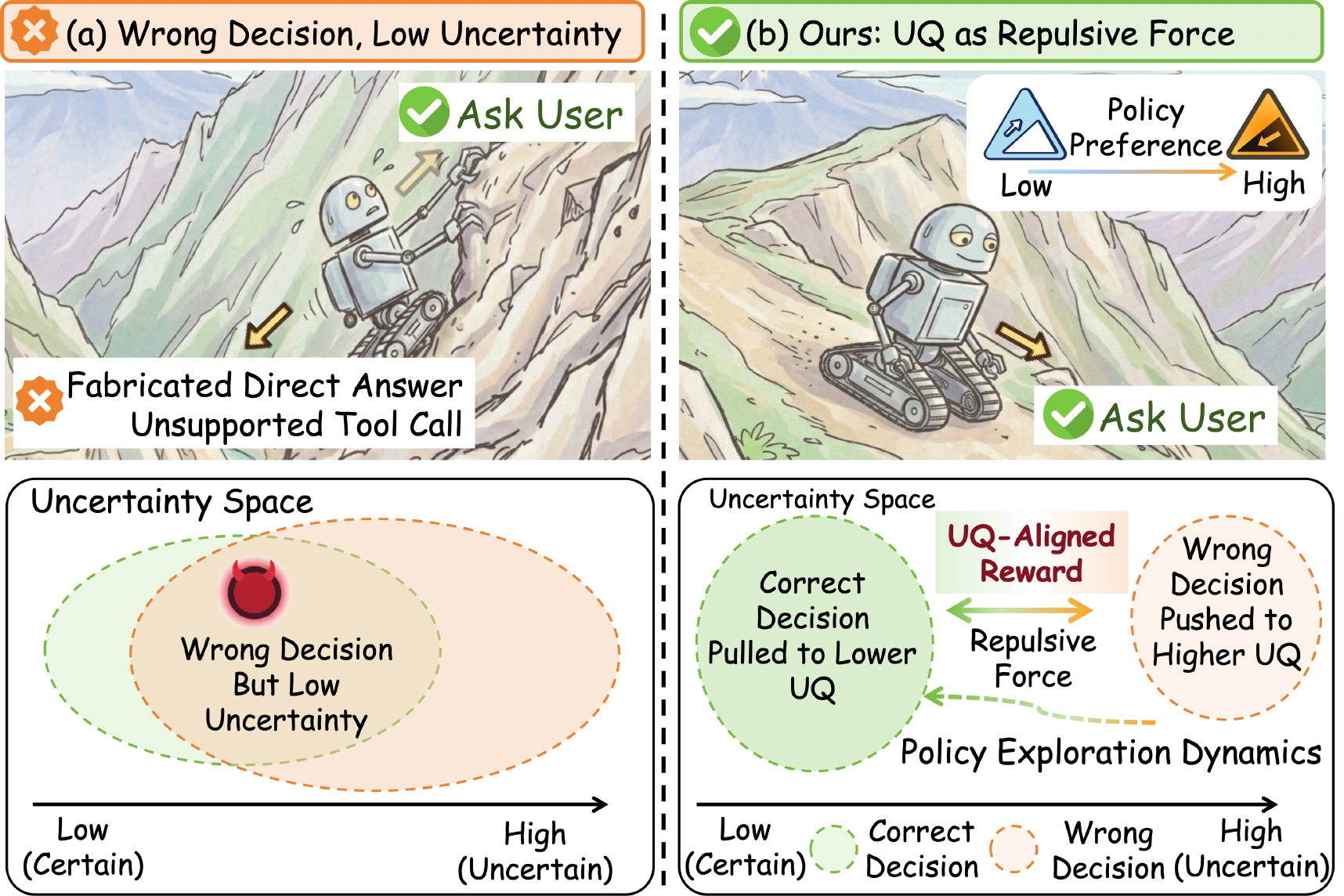}
    \caption{Comparison between tool-calling decision failure in wrong decision but low uncertainty, and our \sys~ solution.}
    \label{fig:1-intro}
\end{figure}

Recent studies address tool-calling decision failures by extracting internal uncertainty signals or structured reasoning patterns during inference \citep{zhang2026agentic,stoisser2025agentsknowdontknow}. However, they primarily focus on post-hoc intervention in inference, instead of improving the agent's intrinsic decision capability through policy optimization. Research efforts by \citet{ross2025when2call,suri2025structured} mitigate such failures through reinforcement learning (RL) for tool-calling decisions, typically relying on coarse-grained reward formulations from decision classification or rule-based checklists. Nevertheless, these approaches lack a quantitative analysis of the policy exploration dynamics underlying tool-calling decision optimization, leaving substantial room for improvement.

From the perspective of \textbf{U}ncertainty \textbf{Q}uantification (UQ) of model responses, we observe that existing RL-based approaches for optimizing tool-calling decisions tend to increase confidence in sampled high-reward actions, while unintentionally weakening the uncertainty separation between correct and incorrect decisions. As illustrated in Fig.~\ref{fig:1-intro}(a), unsupported tool calls and unjustified direct answers increasingly overlap with low-uncertainty regions after decision-oriented RL optimization, impairing the model's original calibration that higher uncertainty should indicate a greater likelihood of incorrect decisions.
Take Qwen3-4B-Thinking \citep{yang2025qwen3} as an example: the proportion of ``Wrong Decision But Low Uncertainty'' cases rises from $34.50\%$ to $70.21\%$ after RL optimization for tool-calling decisions. This observation suggests that existing decision-oriented objectives primarily optimize action correctness without preserving the uncertainty structure underlying those decisions. Consequently, overconfident but incorrect actions receive limited optimization pressure, reducing policy exploration toward more reliable alternatives.

To address this issue, we propose \textbf{T}ool-calling decision \textbf{R}eward with \textbf{U}ncertainty-\textbf{S}eparated post-\textbf{T}raining (\sys), a framework that leverages \textit{UQ as a repulsive force} within reward, rather than using uncertainty merely as a post-hoc diagnostic. Specifically, we introduce a UQ-aligned reward that jointly models action correctness and a certainty margin derived from the uncertainty gap between negative and ground-truth decisions. As shown in Fig.~\ref{fig:1-intro}(b), our reward encourages the model to assign lower uncertainty to correct decisions while maintaining comparatively higher uncertainty for incorrect or counterfactual decisions. This mechanism promotes exploration away from uncertain or unreliable decisions and provides a stronger optimization signal for policy updates.

Furthermore, we extend this training paradigm from standalone decision instances to multi-turn agent trajectories. Instead of exhaustively relabeling entire trajectories, we annotate lightweight key-turn decision points, enabling unified post-training for both trajectory-level task success and turn-level tool-calling calibration. The trajectory-level outcome reward supervises overall task completion and tool execution quality, while our turn-level UQ-aligned reward explicitly calibrates the timing and appropriateness of tool-calling decisions.

Empirically, \sys~ yields substantial improvements across diverse tool-use benchmarks while preserving the uncertainty structure underlying agent decisions. In turn-level tool-calling decision optimization, \sys~ improves over 11\% on the When2Call task accuracy, simultaneously strengthening the performance in complex multi-turn interactions and tool-use trajectories. Cooperated with the trajectory-level tool-calling post-training, \sys~ outperforms 6.33\% on BFCL-V4 and 7.07\% on ToolSandbox. The gains are particularly pronounced in challenging decision-intensive scenarios, including Multi-Turn and Irrelevance on BFCL-V4, as well as scenarios such as multiple user turns and distraction tools in ToolSandbox. Overall, \sys~ jointly optimizes overall tool execution quality and turn-level decision appropriateness of tool use, providing stronger optimization signals for policy learning and enabling more reliable multi-turn agent behavior.

\section{Related Work}

\paragraph{Uncertainty quantification for language agents.}
UQ has been widely studied for estimating the reliability of language model outputs \citep{kadavath2022language, lin2022teaching, zhou2024out}. Early work mainly focuses on output-level uncertainty through model probabilities, verbalized confidence, and sampling-based consistency signals \citep{kadavath2022language,lin2022teaching,manakul2023selfcheckgpt,kuhn2023semantic, manakul2023selfcheckgpt}. More recently, researchers argue that uncertainty in agentic systems extends beyond final responses to intermediate actions, environmental observations, and multi-step trajectories \citep{kirchhof2025position,duan2025uprop,oh2026uncertainty,zhang2026agentic,stoisser2025agentsknowdontknow, lymperopoulos2025tools}. Several approaches further exploit uncertainty to regulate agent behavior, such as triggering clarification under ambiguous instructions, controlling memory and reflection, or guiding exploration through structured rewards \citep{suri2025structured,zhang2026agentic,zhang2026selaur}. Different from prior work that primarily treats uncertainty as a post-hoc signal for diagnosis or behavior control, we explicitly integrate uncertainty into policy optimization and align decision correctness with model certainty through uncertainty-aware rewards.

\begin{figure*}
    \centering
    \includegraphics[width=0.98\linewidth]{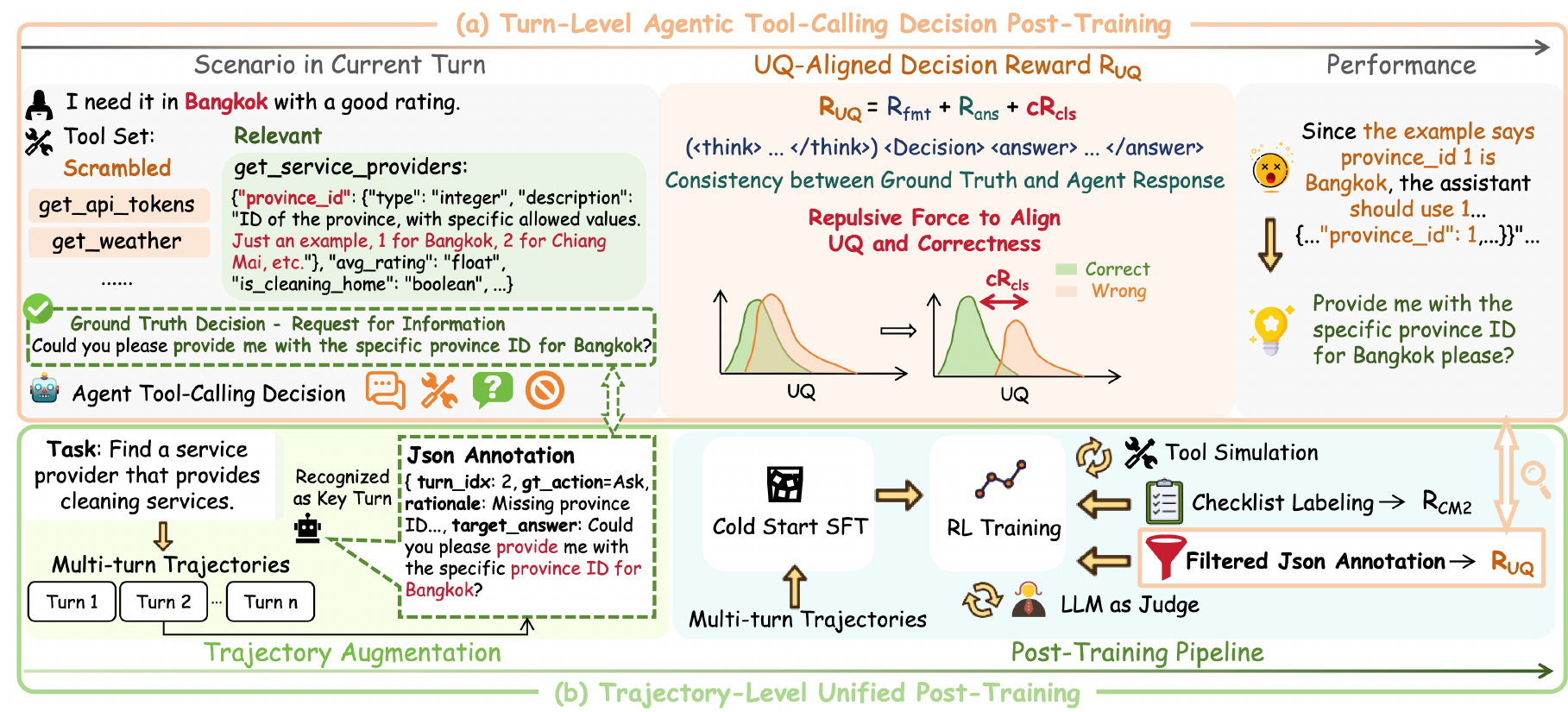}
    \caption{The overview of \sys. It consists of two components: (a) A turn-level UQ-aligned reward uses uncertainty as a repulsive signal to align decision correctness and confidence. (b) A trajectory-level unified post-training framework augments key turns with decision annotations and integrates $R_{\text{UQ}}$ with task-level rewards for joint optimization of execution quality and tool-calling decisions.}
    \label{fig:2-method}
\end{figure*}

\paragraph{Tool-calling decision learning and optimization.}
Tool-augmented LLMs have been extensively studied for tool selection, argument generation, and large-scale API utilization \citep{schick2023toolformer,qin2024toolllm,patil2025bfcl,lu2025toolsandbox}. Beyond execution correctness, recent work increasingly focuses on the preceding decision process: whether an agent should invoke tools, ask follow-up questions, answer directly, or abstain under current observations. Benchmarks and analyses such as When2Call and related studies isolate this decision layer and reveal common failures, including unnecessary tool invocation and hallucinated direct answers \citep{ross2025when2call,wang2025learning,wu2026tocall,sun2026when2tool}. Concurrent work further explores RL-based optimization of tool-calling behavior using calibrated rewards or decision supervision \citep{suri2025structured,modecrua2026multiturn,zhong2026rcgrpo}. In contrast to these approaches that mainly optimize decision correctness, our work studies how RL reshapes the uncertainty structure of tool-calling policies and leverages it to improve both decision quality and calibration.
\section{\sys~ Method}
\label{sec:method}

\subsection{Problem Formulation}
\label{sec:method_formulation}
Given the state $\mathcal{S}$, the next-action decision is a four-way agentic action space \citep{ross2025when2call}:
\begin{equation}
\small 
    \mathcal{A}=\{\textsc{Direct}, \textsc{Tool}, \textsc{Ask}, \textsc{Unable}\},
\end{equation}
where \textsc{Direct} means direct answer, \textsc{Tool} means invoking an external tool, \textsc{Ask} means requesting missing information from the user, and \textsc{Unable} means admitting that the request can not be answered.
Specifically, at decision turn $t$, the agent state is $s_t=(h_t,\mathcal{T})$, where $h_t=(o_1,a_1,...,o_t,a_t)$ denotes the observed trajectory, and $\mathcal{T}$ denotes the available tool set with schemas. A policy $\pi_\theta: \mathcal{S}\xrightarrow{}\mathcal{Z}$ maps from state to agent response, generating a structured response $z_t=(a_t,y_t)$, where $a_t\in\mathcal{A}$ and $y_t\in \mathcal{Y}$, where $\mathcal{Y}$ is the set of surface realizations. For \textsc{Tool}, $y_t$ is a tool-call payload with a tool name and arguments; otherwise, $y_t$ is natural language. The environment updates the state after tool execution or user interaction.

Each decision point provides a ground-truth pair $z_t^\star=(a_t^\star,y_t^\star)$. A tool-calling decision failure occurs when the selected action is unsupported by the current state, \textit{i.e.} $a_t^\star\in\mathcal{A}/\{\textsc{Tool}\}, a_t=\textsc{Tool}$ or $a_t^\star=\textsc{Tool}, a_t=\textsc{Direct}$.
Thus, the training objective is not only to increase the probability of correct final trajectories, but also to calibrate the policy over intermediate decisions:
\begin{equation}
    \max_{\theta}~\pi_\theta(a_t^\star,y_t^\star \mid s_t),~ \min_{\theta}~\pi_\theta(a_t^-,y_t^- \mid s_t)
\end{equation}
for unsupported or counterfactual decisions $z_t^-=(a_t^-,y_t^-)$. In our experiments, this policy is optimized with GRPO post-training. For each state $s$, the current model samples a group of responses $\mathcal{Z}^G=\{z_i\}_{i=1}^{G}$, and the policy is updated according to their relative rewards.

\subsection{Turn-level UQ-Aligned Decision Reward}
\label{sec:method_uq_reward}

\begin{figure*}[t]
    \centering
    \includegraphics[width=\linewidth]{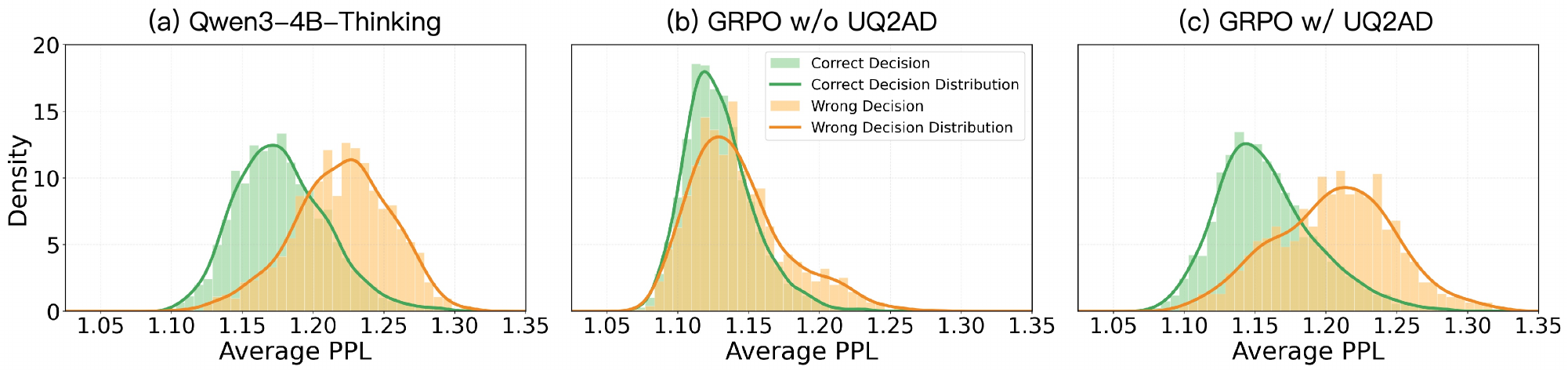}
    \caption{Uncertainty calibration of tool-calling decisions on When2Call. Green denotes correct decisions, and orange denotes wrong decisions. Lower PPL means higher certainty. Compared with Qwen3-4B-Thinking and decision GRPO training without \sys, training with \sys~ yields a clearer separation between correct and wrong decisions, assigning lower PPL to correct decisions and higher PPL to wrong ones.}
    \label{fig:3-motiv}
\end{figure*}

Directly rewarding the selected action can not make the agent's uncertainty better calibrated, as shown in Fig.~\ref{fig:3-motiv}. So we use UQ as the repulsive force in reward as illustrated in Fig.~\ref{fig:2-method}(a). We instantiate uncertainty with sequence perplexity \citep{kuhn2023semantic}. Given a prompt state $s$ and a candidate response $z=(a,y)$, the perplexity is
\begin{equation}
\small
    \mathrm{PPL}_{\theta}(z\mid s)
    =
    \exp\left(
    -\frac{1}{|z|}
    \sum_{j=1}^{|z|}
    \log p_{\theta}(z_{(j)}\mid s)
    \right),
\end{equation}
where $z_{(j)}$ is the j-th sequence in $z$, $p_{\theta}$ calculates the average value of tokens in $z_{(j)}$.

For each decision point, we compare the perplexity of the ground-truth decision with that of a negative decision:
\begin{equation}
\small
    m(s) = \mathbb{E}_{\mathcal{Z}^G}(\mathrm{PPL}_{\theta}(z^- \mid s))
    -
    \mathbb{E}_{\mathcal{Z}^G}(\mathrm{PPL}_{\theta}(z^\star \mid s)),
\end{equation}
where {\small$\mathbb{E}_{\mathcal{Z}^G}(\mathrm{PPL}_{\theta}(z^- \mid s))=1$} if {\small$\{(a^*,\cdot)\in\mathcal{Z}^G\}=\emptyset$}. The margin is converted into a bounded certainty coefficient,
\begin{equation}
\small
    c(s) = \sigma\left(\frac{m(s)}{\tau}\right),
\end{equation}
where $\tau=0.1$ is a temperature and $\sigma(\cdot)$ is the sigmoid function. A large margin means that the model assigns lower perplexity to the correct decision than to the negative decision. Conversely, a small or negative margin indicates that uncertainty is not aligned with decision correctness.

The reward combines format validity, decision correctness, and the UQ margin:
\begin{equation}
\small
    R_{\mathrm{UQ}}(z)
    =
    R_{\mathrm{fmt}}(z)
    +
    R_{\mathrm{ans}}(z,z^\star)
    +
    c R_{\mathrm{cls}}(a,a^\star).
    \label{eq:uq_reward}
\end{equation}
Here, $R_{\mathrm{fmt}}$ rewards structured output format {\small \texttt{(<think> ... </think>)<$\mathcal{A}$><answer>$\mathcal{Y}$</answer>}} and internal consistency between the suggested action $a$ and final answer $y$. n trajectory-level unified training, $R_{\text{fmt}}$ is set to 0. $R_{\mathrm{ans}}(z,z^\star)$ rewards consistency between $z$ and $z^*$. $R_{\mathrm{cls}}$ rewards action correctness and certainty coefficient $c$ itself: 
\begin{equation}
\small
\begin{aligned}
    & R_{\mathrm{cls}}(a,a^\star) = 1 + \\
    & \begin{cases}
        2&, a^*=a, \\
        1&, \{a,a^*\}=\{\textsc{Direct}, \textsc{Tool}\}~\text{or}~\{\textsc{Ask}, \textsc{Unable}\}, \\
        0&, \text{otherwise}.
    \end{cases}
\end{aligned}
\end{equation}
Exact action match receives the highest $R_{\mathrm{cls}}$, while a weaker partial reward is assigned when the prediction preserves the coarse decision direction between action and non-execution. 
$R_{\mathrm{UQ}}$ can be obtained without a LLM judger if using structured output format {\small \texttt{(<think> ... </think>)<$\mathcal{A}$><answer>$\mathcal{Y}$</answer>}}; otherwise, it needs a lightweight LLM judger to judge the action $a\in \mathcal{A}$.
Overall, $cR_{\text{cls}}$ gives a repulsive force in $R_{UQ}$, to push wrong decisions to higher UQ and pull correct decisions to lower UQ, driving policy expands the exploration space in cases of incorrect decisions, and makes it easier to sample the right ones.

\subsection{Unified Post-Training for Trajectory-Level Performance and Turn-Level Decision}
\label{sec:method_trajectory_aug}

Standalone decision examples provide clean supervision for when an agent should answer, call tools, ask the user, or stop. However, realistic agent failures often arise inside multi-turn trajectories, where an early unsupported decision can corrupt later states. Therefore, it is of great significance to unify the classical task performance-driven rewards and the next-action decision rewards for RL training. To realize this unification, we augment CM2 trajectories \citep{zhang2026cm2reinforcementlearningchecklist} with lightweight tool-call decision annotations at key turns, and plug in our $R_{\mathrm{UQ}}$ with the CM2 checklist reward $R_{\text{CM2}}$.

\paragraph{Trajectory augmentation with tool-call decision annotation.}
\label{sec:data_construction}
As shown in Fig.~\ref{fig:2-method}(b), we construct trajectory-level decision supervision from trajectories without relabeling the full conversation, only augmenting each trajectory with a small set of key-turn decision annotations.

A labeling model reads the full payload and selects no more than 2 decision-critical turns where a next-action $\mathcal{A}$ signal is useful. Each selected turn receives exactly one label from the action space $\mathcal{A}$. To reduce the annotation bias of trajectories toward tool execution \textsc{Tool}, the prompt for the labeling model includes running label-count statistics and requests to prefer underrepresented valid actions when multiple turns are equally useful.

The annotation output is a strict JSON object. Each annotation has the following schema
\begin{equation}
\small
\begin{aligned}
\texttt{annotation}=\{& \\
&\texttt{turn\_idx}: \text{key turn idx}, \\
&\texttt{gt\_action}=a^* \in \mathcal{A}, \\
&\texttt{rationale}: \text{Reason for } \texttt{gt\_action}, \\
&\texttt{target\_answer}=y^*\in \mathcal{Y} \\
\}
\end{aligned}
\end{equation}
which is used by the mixed reward manager during GRPO post-training. For each annotated turn, the trajectory prefix up to the user message defines the state $s_t$, the annotated label defines $a_t^\star$, and the annotated target answer defines $y_t^\star$.
\paragraph{Reward combination for unified RL.} 
To unify the next-action decision with the task completion and accuracy aggregation in one post-training process, trajectories are optimized with a mixed reward as shown in Fig.~\ref{fig:2-method}(b). The original checklist reward $R_{\text{CM2}}$ continues to supervise task completion and tool execution quality over the full trajectory. Our turn-level decision reward $R_{\text{UQ}}$ is added sparsely at the annotated turn end positions:
\begin{equation}
\small
    R
    =
    R_{\text{CM2}}
    +
    \sum_{t\in\mathcal{K}}
    R_{\mathrm{UQ}}(z_t),
\end{equation}
where $\mathcal{K}$ is the set of annotated key turns. This design preserves the original CM2 objective for execution quality while adding a targeted next-action decision signal for tool-call timing and action appropriateness.
\sys~ yields a lightweight trajectory-level training signal without relabeling entire conversations or redesigning the performance-driven rewards, realizing the unified training for task performance, efficiency for tool-calling, and tool-calling hallucination mitigation.

\section{Experiments}
\begin{table*}[t]
    \centering
    \small
    \caption{Performance comparison of When2Call. FDAR is the False Direct Answer Rate. \textbf{Boldface} and \underline{underlining} denote the best and second-best results, respectively, within each corresponding baseline group. 
    The transparency intensity of \hl{green-colored cells} is (Performance of \sys) - (Performance of The Baseline with same ``Training'').}
    \label{tab: when2call}
    \tabcolsep=0.5cm
    \resizebox{\linewidth}{!}{
    \begin{tabular}{l|cc|cccc}
    \specialrule{0.1em}{0em}{0em}
        \textbf{Model / Method} & \textbf{Model Base} & \textbf{Training} & \textbf{Acc Norm}$\uparrow$ & \textbf{F1}$\uparrow$ & \textbf{Tool Hall}$\downarrow$ & \textbf{FDAR}$\downarrow$ \\ \hline
        \multicolumn{7}{l}{{\cellcolor{gray!6}\textit{Closed-source Baselines}}} \\ 
        MiniMax-M2.5 & -& -& 23.93 & 24.94 & 24.03 & 20.48\\ 
        GPT-4o-mini & -& -&57.64 & 60.26 & 59.30 & 11.61\\ 
        Claude-Sonnet-4 & -& -&80.53 & 81.85 & 10.08 & 5.86 \\ \hline
        \multicolumn{7}{l}{{\cellcolor{gray!6}\textit{Open-source Baselines}}}  \\ 
        235B-A22B-Instruct & -& -&42.33 & 32.62 & 82.95 & 4.63  \\ 
        30B-A3B-Instruct & -& -&35.10 & 17.75 & 94.57 & 1.34  \\ 
        4B-Thinking & -& -&69.36 & 72.77 & 41.47 & 17.11 \\ 
        8B-Thinking & - & - & 45.19 & 50.30 & 14.73 & 32.78 \\ \hline
        \multicolumn{7}{l}{{\cellcolor{gray!6}\textit{From Qwen3-4B-Thinking}}} \\
        
        AUQ & 4B-Thinking & Training-free &56.35 & 60.32 & 45.71 & 22.98 \\ 
        SAGE & 4B-Thinking & Training-free & \underline{73.36} & 72.36 & \textbf{0.00} & \underline{8.89} \\ 
        GRPO & 4B-Thinking & Turn-level & 72.46 & \underline{76.06} & \underline{5.73} & 24.76 \\
        \textbf{Turn-level \sys} & 4B-Thinking & Turn-level  & {\cellcolor{green!8.37}\textbf{80.83}} & {\cellcolor{green!6.78}\textbf{82.84}} & 17.83 & {\cellcolor{green!19.69}\textbf{5.07}} \\ \hline
        \multicolumn{7}{l}{{\cellcolor{gray!6}\textit{From Qwen3-8B-Thinking}}} \\        
        AUQ & 8B-Thinking & Training-free & 34.34 & 37.91 & \underline{37.21} & 36.99 \\ 
        SAGE & 8B-Thinking & Training-free & \textbf{71.87} & \textbf{71.57} & \textbf{0.00} & 11.04 \\
        \multicolumn{7}{l}{{\cellcolor{gray!6}\textit{From Qwen3-8B-Base}}} \\
        Traj.-level CM2 & 8B-Base & Traj.-level &43.75 & 36.94 & 75.43 & \underline{9.34} \\ 
        \textbf{Traj.-level \sys} & 8B-Base & Traj.-level & {\cellcolor{green!18.57}\underline{62.32}} & {\cellcolor{green!23.68}\underline{60.62}} & {\cellcolor{green!25.56}49.87} & {\cellcolor{green!0.95}\textbf{8.39}} \\ \specialrule{0.1em}{0em}{0em}
    \end{tabular}
}
\end{table*}
\subsection{Setup}
\paragraph{Benchmarks.}
We conduct experiments on three benchmarks, namely When2Call \citep{ross2025when2call} for turn-level when to call training and evaluation, and tool-use benchmarks ToolSandbox \citep{lu2025toolsandbox} and BFCL-V4 \citep{patil2025bfcl} to evaluate the multi-turn overall performance.

\paragraph{Baselines.}
Closed models MiniMax-M2.5 \citep{minimax}, GPT-4o-mini \citep{hurst2024gpt}, and open-source models Qwen3-235B-A22B, Qwen3-4B-Thinking, and Qwen3-8B-Thinking \citep{yang2025qwen3} are tested on three benchmarks for performance comparison of our post-trained model with \sys. Moreover, we compare two representative UQ for agentic tool-calling baselines, AUQ \citep{zhang2026agentic} and SAGE-Agent \citep{suri2025structured}, all implemented on Qwen3-4B-Thinking and Qwen3-8B-Thinking. Note that we do not report the results of the vanilla base model and training-free methods for Qwen3-8B-Base, since it is not instruction-tuned under this evaluation protocol \citep{zhang2026cm2reinforcementlearningchecklist}. Instead, we provide the corresponding training-free experiments on Qwen3-8B-Thinking as informative references. For turn-level and trajectory-level post-training, GRPO only by the tool-calling decision, and CM2 \citep{zhang2026cm2reinforcementlearningchecklist}, from the same model checkpoint as \sys, serve as training-needed baselines. See Appendix~\ref{app: Baselines} for further details of benchmarks and baselines.

\paragraph{Training setups.}
We select Qwen3-4B-Thinking and Qwen3-8B-Base as the agent backbone models \citep{yang2025qwen3} for group relative policy optimization (GRPO) post-training \citep{shao2024deepseekmath}. \textit{Qwen3-4B-Thinking} is trained on When2Call training dataset to validate the efficiency of calibrating \textit{turn-level tool-calling decision} only, while \textit{Qwen3-8B-Base} sees \textit{a complete trajectory-level post-training}, including cold start SFT and the unified RL, following \citep{zhang2026cm2reinforcementlearningchecklist}. In trajectory augmentation, we use Qwen3-235B-A22B-Instruct-2507 \citep{yang2025qwen3} as the labeler. In unified RL training, Qwen3-30B-A3B-Instruct-2507 \citep{yang2025qwen3} serves as an LLM judger. Prompts and more settings are in Appendix~\ref{app: Prompts and Settings}.

\subsection{\sys~ Successfully Aligns Correctness and Uncertainty}
\begin{table*}[t]
    \centering
    \caption{Performance comparison of BFCL-V4.}
    \label{tab: bfcl}
    \small
    \tabcolsep=0.4cm
    \resizebox{\linewidth}{!}{
    \begin{tabular}{l|c|cccccc}
    \specialrule{0.1em}{0em}{0em}
        \textbf{Model / Method} & \textbf{Overall Score} & \textbf{Web Search} & \textbf{Memory} & \textbf{Multi-Turn} & \textbf{Live} & \textbf{Non-Live} & \textbf{Irrelevance} \\ \hline
        \multicolumn{8}{l}{{\cellcolor{gray!6}\textit{Open-source Baselines}}}\\
        235B-A22B-Instruct & 53.54 & 44.50 & 27.10 & 50.12 & 83.20 & 90.10 & 82.38 \\ 
        30B-A3B-Instruct & 41.00 & 22.50 & 17.63 & 30.00 & 77.94 & 85.77 & 79.90 \\ 
        4B-Thinking & 38.61 & 6.00 & 12.37 & 52.00 & 71.06 & 38.02 & 79.23 \\ 
        8B-Thinking & 41.58 & 12.00 & 14.62 & 37.75 & 80.53 & 87.58 & 79.07 \\ \hline
        \multicolumn{8}{l}{{\cellcolor{gray!6}\textit{From Qwen3-4B-Thinking}}}\\        
        AUQ & 35.21 & 3.00 & 20.43 & 23.50 & 62.40 & 77.08 & 81.37 \\ 
        SAGE & \underline{41.16} & 3.50 & 13.02 & 41.50 & 80.49 & 86.71 & 79.23 \\ 
        Turn-level GRPO & 38.98 & 3.00 & 24.52 & 22.75 & 86.33 & 80.61 & 82.38 \\
        \textbf{Turn-level \sys}  & {\cellcolor{green!9.06}\textbf{48.04}} & 3.00 & 24.95 & 49.62 & 84.09 & 89.23 & 84.84 \\ \hline
        \multicolumn{8}{l}{{\cellcolor{gray!6}\textit{From Qwen3-8B-Thinking}}}\\
        
        AUQ & 29.08 & 14.00 & 15.70 & 10.00 & 75.42 & 84.46 & 40.14 \\ 
        SAGE & \underline{42.99} & 9.50 & 21.72 & 36.37 & 80.83 & 88.15 & 79.60 \\ 
        \multicolumn{8}{l}{{\cellcolor{gray!6}\textit{From Qwen3-8B-Base}}}\\
        Traj.-level CM2 & 38.15 & 22.00 & 21.94 & 22.50 & 74.58 & 80.12 & 71.46 \\ 
        \textbf{Traj.-level \sys} & {\cellcolor{green!6.33}\textbf{44.48}} & 27.00 & 30.57 & 29.88 & 75.57 & 82.27 & 79.33 \\ \specialrule{0.1em}{0em}{0em}
    \end{tabular}
}
\end{table*}

\begin{table*}[t]
\centering
\caption{Performance comparison of ToolSandBox benchmark across scenario categories and tool augmentations.}
\label{tab:main_results}
\renewcommand{\arraystretch}{1.25}
\setlength{\tabcolsep}{4pt}
\resizebox{\textwidth}{!}{
\begin{tabular}{l | c | c c c c c c c | c c c c c c c c}
\specialrule{0.1em}{0em}{0em}
\multirow{2}{*}{\textbf{Model / Method}} &
\multirow{2}{*}{\textbf{Overall Score}} &
\multicolumn{7}{c|}{\textbf{Scenario Categories}} &
\multicolumn{8}{c}{\textbf{Tool Augmentations}} \\
\cmidrule(lr){3-9} \cmidrule(lr){10-17}
& & \textbf{STC} & \textbf{MTC} & \textbf{SUT} & \textbf{MUT} & \textbf{SD} & \textbf{C} & \textbf{II}
& \textbf{0-DT} & \textbf{3-DT} & \textbf{10-DT} & \textbf{AT} & \textbf{TNS} & \textbf{TDS} & \textbf{ADS} & \textbf{ATS} \\
\midrule

\multicolumn{17}{l}{{\cellcolor{gray!6}\textit{Open-source Baselines}}}\\

235B-A22B-Instruct
& 69.88 & 62.46 & 62.68 & 63.85 & 59.48 & 62.68 & 57.63 & 90.98
& 73.84 & 73.44 & 74.98 & 70.75 & 74.85 & 70.99 & 74.73 & 74.79 \\

30B-A3B-Instruct
& 61.57 & 62.08 & 60.07 & 62.88 & 54.09 & 62.34 & 57.24 & 58.45
& 64.68 & 62.12 & 62.25 & 61.72 & 63.61 & 64.94 & 63.54 & 63.55 \\ 

4B-Thinking
& 52.89 & 57.68 & 38.09 & 43.15 & 38.18 & 53.03 & 28.71 & 90.82
& 58.71 & 54.75 & 53.21 & 52.61 & 53.82 & 53.64 & 56.40 & 60.54 \\ 

8B-Thinking
& 59.74 & 64.32 & 44.79 & 50.16 & 44.06 & 61.55 & 35.76 & 94.93
& 59.05 & 62.95 & 65.13 & 60.74 & 62.35 & 61.08 & 65.23 & 64.00 \\ \hline

\multicolumn{17}{l}{{\cellcolor{gray!6}\textit{From Qwen3-4B-Thinking}}}\\

AUQ
& \underline{55.47} & 48.23 & 47.92 & 49.20 & 44.79 & 51.59 & 41.71 & 84.87
& 56.12 & 58.09 & 60.55 & 56.98 & 60.41 & 59.03 & 55.51 & 56.99 \\

SAGE
& 52.54 & 48.22 & 46.93 & 49.14 & 42.04 & 54.38 & 40.99 & 71.06
& 50.14 & 56.03 & 54.33 & 51.60 & 60.46 & 51.49 & 55.90 & 55.45 \\

Turn-level GRPO & 39.37 & 46.92 & 21.15 & 28.67 & 19.04 & 23.37 & 19.05 & 95.98 & 43.36 & 45.57 & 40.93 & 38.86 & 44.28 & 37.90 & 41.82 & 43.71 \\
\textbf{Turn-level \sys}
& {\cellcolor{green!6.98}\textbf{56.35}} & 51.18 & 46.26 & 52.83 & 37.68 & 55.59 & 42.27 & 94.44
& 57.42 & 58.53 & 59.05 & 58.68 & 57.66 & 56.90 & 58.02 & 58.77 \\ \hline

\multicolumn{17}{l}{{\cellcolor{gray!6}\textit{From Qwen3-8B-Thinking}}}\\

AUQ
& \underline{64.08} & 56.46 & 56.68 & 57.85 & 53.48 & 56.68 & 51.63 & 87.98
& 67.84 & 67.44 & 68.98 & 64.75 & 68.85 & 64.99 & 68.73 & 68.79 \\

SAGE
& 59.72 & 54.85 & 51.56 & 53.95 & 47.56 & 59.41 & 43.68 & 83.48
& 59.70 & 65.72 & 60.27 & 58.29 & 62.97 & 66.61 & 62.22 & 65.52 \\

\multicolumn{17}{l}{{\cellcolor{gray!6}\textit{From Qwen3-8B-Base}}}\\

Traj.-level CM2
& 61.21 & 70.57 & 58.55 & 64.22 & 51.93 & 56.64 & 53.70 & 55.60
& 60.12 & 65.04 & 65.02 & 65.03 & 61.80 & 61.30 & 64.14 & 64.51 \\

\textbf{Traj.-level \sys}
& {\cellcolor{green!7.07}\textbf{68.28}} & 68.66 & 59.70 & 63.10 & 56.93 & 61.58 & 52.04 & 90.70
& 73.01 & 76.41 & 74.94 & 69.16 & 66.74 & 68.77 & 71.26 & 71.24 \\

\specialrule{0.1em}{0em}{0em}
\end{tabular}
}
\end{table*}

\begin{figure*}[t]
    \centering
    \includegraphics[width=\linewidth]{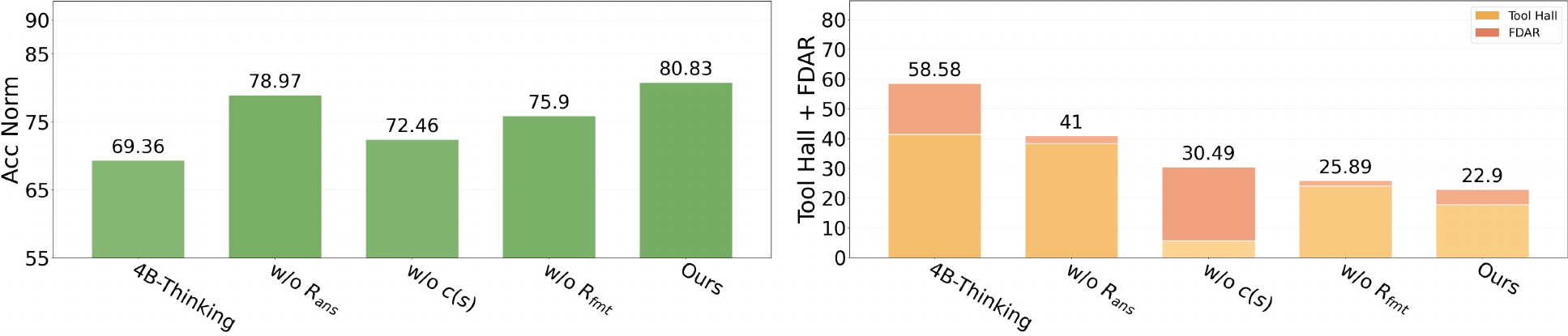}
    \caption{Ablation study of $R_{\text{UQ}}$ on When2Call. Tool Hall + FDAR measures the overall tool-calling hallucination. The setting w/o $c(s)$ represents $c(s)=1$, \textit{i.e.} $R_{\text{UQ}}=R_{\mathrm{fmt}}+R_{\mathrm{ans}}+R_{\mathrm{cls}}$.}
    \label{fig:4-ablation_reward}
    % \vspace{-0.5em}
\end{figure*}

\begin{figure*}[t]
    \centering
    \includegraphics[width=\linewidth]{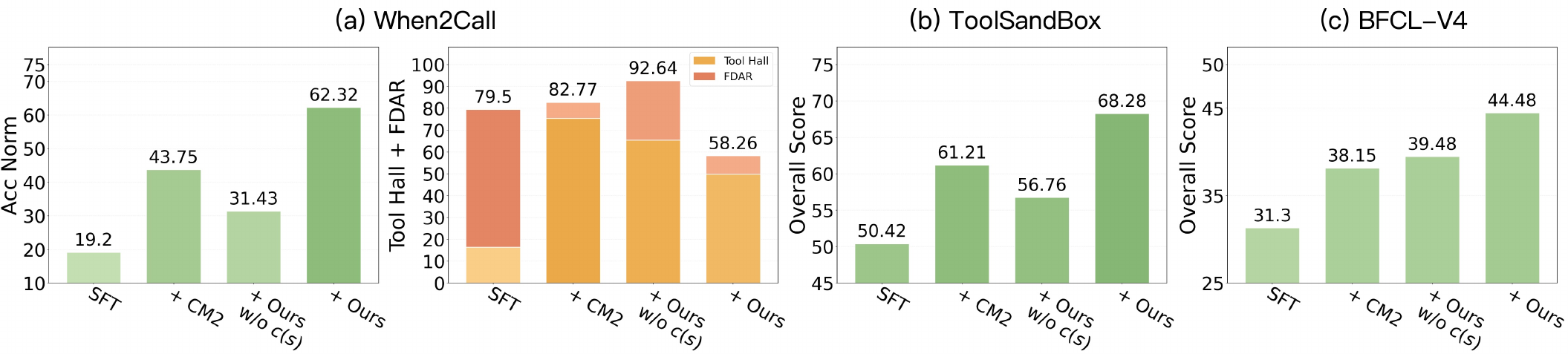}
    \caption{Ablation study results of the unified post-training across three evaluation benchmarks.}
    \label{fig:5-ablation_8b}
    % \vspace{-0.3em}
\end{figure*}

To isolate whether the proposed reward calibrates the tool-calling decision itself, we train on When2Call training set and compare PPL distributions of correct and wrong decisions. As shown in Fig.~\ref{fig:3-motiv}(a), the original Qwen3-4B-Thinking already exhibits a correlation between correctness and uncertainty, where correct decisions tend to have lower PPL than wrong ones, forming two distinct distributions. After GRPO decision training without \sys, however, this separation becomes less reliable in Fig.~\ref{fig:3-motiv}(b). Wrong decisions are also assigned low PPL, indicating that direct optimization can make the model confident in both correct and incorrect decisions. In contrast, training with \sys~ again produces a clearer uncertainty gap depicted in Fig.~\ref{fig:3-motiv}(c). Correct decisions concentrate in the lower-PPL region, while wrong decisions are shifted toward higher PPL. 

We measure the distribution overlap in Fig.~\ref{fig:3-motiv} by 
\begin{equation}
\small
    \text{IoU}=\frac{\text{Correct} \cap \text{Wrong}}{\text{Correct} \cup \text{Wrong}}.
\end{equation}
It yields an initial IoU of 34.50\%, which increases to 70.21\% following GRPO. Conversely, integrating our UQ-Aligned Reward leads to a reversion of IoU to 35.29\% while delivering superior accuracy.
This demonstrates that \sys~ explicitly aligns the model's internal uncertainty with decision correctness, making erroneous tool-calling decisions easier to identify and less likely to be reinforced as confident actions.

\subsection{Performance of \sys~ on Turn-Level Tool-Calling Decision Post-Training}

To evaluate the effectiveness of \sys, we conduct turn-level GRPO and evaluation on Qwen3-4B-Thinking, where the results demonstrate substantial performance gains across all benchmarks.

First, on \textbf{When2Call} test set (Table~\ref{tab: when2call}), \sys~ achieves the highest Acc Norm compared to various baseline methods, outperforming the runner-up by 7.47\%, while yielding the lowest False Direct Answer Rate (FDAR) of 5.07\%. 
Compared to the vanilla base model Qwen3-4B-Thinking, \sys~ improves the Acc Norm by 11.47\%. 
When compared directly against turn-level GRPO training, \sys~ not only achieves an 8.37\% absolute improvement in Acc Norm but also reduces the overall hallucination metric, defined as a sum of Tool Hallucination and FDAR, from GRPO's 30.49\% down to 22.90\%. 
Overall, \sys~ surpasses most open-source and closed-source models, reaching a level comparable to the top-tier closed-source Claude-Sonnet-4.

Crucially, we discover that \textit{optimizing tool-call decisions solely at turn level can directly catalyze and boost the generalized performance in complex multi-turn interactions and tool-use trajectories}. 
On \textbf{BFCL-V4} in Table~\ref{tab: bfcl}, \sys~ delivers an Overall Score of 48.04\%, which represents a substantial 9.43\% leap over the base model, completely dominates all baseline methods, and notably exceeds the performance of the much larger 30B-A3B-Instruct (41.00\%). 
Specifically, \sys~ exhibits a compelling advantage in dimensions like Multi-Turn (49.62\%) and Irrelevance (84.84\%), proving its high precision in handling multi-turn tool-calling and detecting irrelevant tools. 

Similarly, on \textbf{ToolSandBox} in Table~\ref{tab:main_results}, \sys~ establishes a state-of-the-art performance within its group, outperforming the base model by 3.46\% and scoring over turn-level GRPO with a 16.98\% margin. In contrast to \sys, while turn-level GRPO has the highest performance in detecting Insufficient Information (II), it drives the agent towards a conservative tool-calling strategy, which affects the tool-calling and task completion capability. Furthermore, when subjected to various tool-level perturbations 3-DT and 10-DT as well as specialized robust evaluations (TNS, TDS, ADS, and ATS), \sys~ maintains strong score stability and resilience. This firmly underscores the powerful cross-task benefits that precise turn-level decision training provides for long-dependency tool-chain workflows.

\subsection{Performance of \sys~ on Trajectory-Level Unified Post-Training}

On \textbf{When2Call} in Table~\ref{tab: when2call}, \sys~ demonstrates a commanding superiority over the trajectory-level RL baseline CM2. 
Specifically, \sys~ advances the Acc Norm from 43.75\% to 62.32\% and the F1 score from 36.94\% to 60.62\%, marking a substantial leap in execution accuracy.

Furthermore, the experimental empirical results on multi-turn benchmarks confirm that \sys, which applies GRPO on trajectories augmented with key turn decision annotations, utilizes tools more accurately and effectively. 
On \textbf{BFCL-V4} in Table~\ref{tab: bfcl}, \sys~ establishes an Overall Score of 44.48\%, significantly outperforming Traj.-level CM2 (38.15\%) and surpassing 30B-A3B-Instruct (41.00\%). 
This outstanding performance is consistently mirrored in scenarios requiring higher agent capabilities, such as Web Search, Memory, and Multi-Turn dimensions. 
Similarly, on \textbf{ToolSandBox} in Table~\ref{tab:main_results}, \sys~ delivers an  Overall Score of 68.28\%, outstripping the trajectory-level GRPO baseline by 7.07\%. 
Remarkably, this score not only leaves the 30B model far behind but also closely approaches the performance of 235B-A22B-Instruct (69.88\%). 
Moreover, it is highlighted that \sys~ achieves an Insufficient Information (II) score of 90.70\%, drastically mitigating the hallucination plagued by the GRPO baseline CM2 (55.60\%).

Importantly, unlike training-free alternatives (\textit{e.g.}, SAGE, AUQ) that often rely on multi-step prompting or trial-and-error reasoning loops, \sys~ internalizes the decision-making logic directly into the model weights, thereby introducing zero additional inference latency or computational overhead during deployment.

\subsection{Ablation Study}

\paragraph{Ablation for $R_{\text{UQ}}$.}
As shown in Fig.~\ref{fig:4-ablation_reward}, all experiments trained with GRPO demonstrate improvements over the vanilla model baseline, while our full method achieves the best overall performance. Among all reward components, removing the uncertainty repulsive reward \textit{i.e.} w/o $c(s)$ leads to the most significant degradation, reducing Acc Norm from 80.83\% to 72.46\%. Furthermore, excluding the answer reward (w/o $R_{\text{ans}}$) or the format reward (w/o $R_{\text{fmt}}$) also results in varying degrees of performance decline, accompanied by lower accuracy and increased hallucination compared with the complete method. Notably, removing $c(s)$ also substantially increases the overall hallucination (Tool Hall + FDAR) from 22.90\% to 30.49\%, indicating that $c(s)$ plays a crucial role in suppressing unsupported tool calls and false direct answers. These results suggest that different reward components contribute complementary effects, where $R_{\text{ans}}$ and $R_{\text{fmt}}$ mainly improve answer quality and output validity, while $c(s)$ is particularly important for reliable tool-calling decisions. 

\paragraph{Ablation for unified post-training.}
As shown in Fig.~\ref{fig:5-ablation_8b}, in trajectory-level unified post-training, removing $c(s)$ also leads to a drastic performance degradation across all evaluated benchmarks, firmly validating that the repulsive force in \sys~ is vital for calibrating tool-call decisions. Compared with SFT and the GRPO baseline without tool-calling decision annotation and a combination of rewards in \sys, the full unified training framework achieves the highest overall scores and the lowest hallucination across the board.
The empirical results consistently demonstrate that every designed module plays an indispensable role, and their unification yields the optimal performance. More experimental result details and case study can be found in Appendix~\ref{app: results} and Appendix~\ref{app: case}.
\section{Conclusions}
In this paper, we propose \sys, an uncertainty-aware reward optimization framework that improves agentic tool-calling decisions in both turn-level and trajectory-level unified post-training. By integrating uncertainty quantification information directly into the reward function, \sys~ aligns model certainty with decision correctness and mitigates hallucinated tool-use behaviors. We further introduce lightweight key-turn annotations to unify turn-level decision calibration with trajectory-level task optimization. Extensive experiments on When2Call, BFCL-V4, and ToolSandbox demonstrate that \sys~ consistently improves tool-calling reliability, multi-turn interaction quality, and overall agent performance without additional inference overhead.

\section*{Limitations}
Despite its effectiveness, several limitations remain for future exploration. First, the current framework mainly relies on perplexity-based uncertainty estimation, while more advanced semantic or trajectory-level uncertainty modeling could further improve calibration. Second, our experiments focus on text-based tool-use benchmarks with predefined action spaces. Extending \sys~ to more dynamic settings, such as embodied agents or open-world tool ecosystems, is an important direction for future work.

\section*{Ethical Considerations}
This work aims to improve the tool-calling decision capability and reliability of LLM-based agents by proposing \sys, an uncertainty-aware reward framework for post-training tool-calling decisions. By aligning decision confidence with correctness, our method improves the turn-level tool-calling decision accuracy and the trajectory-level task performance, while reducing hallucinated or unsupported tool use and mitigating error propagation in multi-turn interactions. All datasets and reproduced baselines used in this work are publicly available,  properly cited, and follow their original licenses and terms of use. While we do not anticipate direct severe societal risks from the proposed method itself, more capable agentic systems may still be misused for generating misleading content or unsafe automated behaviors. Future work should further strengthen safety mechanisms such as content moderation and risk-aware deployment protocols.

{
    \bibliography{custom}

@inproceedings{yao2023react,
    title={ReAct: Synergizing Reasoning and Acting in Language Models},
    author={Yao, Shunyu and Zhao, Jeffrey and Yu, Dian and Du, Nan and Shafran, Izhak and Narasimhan, Karthik and Cao, Yuan},
    booktitle={International Conference on Learning Representations},
    year={2023}
  }

@misc{schick2023toolformer,
      title={Toolformer: Language Models Can Teach Themselves to Use Tools}, 
      author={Timo Schick and Jane Dwivedi-Yu and Roberto Dessì and Roberta Raileanu and Maria Lomeli and Luke Zettlemoyer and Nicola Cancedda and Thomas Scialom},
      year={2023},
      eprint={2302.04761},
      archivePrefix={arXiv},
      primaryClass={cs.CL},
      url={https://arxiv.org/abs/2302.04761}, 
}

@inproceedings{qin2024toolllm,
  title={Toolllm: Facilitating large language models to master 16000+ real-world apis},
  author={Qin, Yujia and Liang, Shihao and Ye, Yining and Zhu, Kunlun and Yan, Lan and Lu, Yaxi and Lin, Yankai and Cong, Xin and Tang, Xiangru and Qian, Bill and others},
  booktitle={International Conference on Learning Representations},
  volume={2024},
  pages={9695--9717},
  year={2024}
}

@inproceedings{ross2025when2call,
  title={When2Call: When (not) to call tools},
  author={Ross, Hayley and Mahabaleshwarkar, Ameya Sunil and Suhara, Yoshi},
  booktitle={Proceedings of the 2025 Conference of the Nations of the Americas Chapter of the Association for Computational Linguistics: Human Language Technologies (Volume 1: Long Papers)},
  pages={3391--3409},
  year={2025}
}

@article{shao2024deepseekmath,
  title={Deepseekmath: Pushing the limits of mathematical reasoning in open language models},
  author={Shao, Zhihong and Wang, Peiyi and Zhu, Qihao and Xu, Runxin and Song, Junxiao and Bi, Xiao and Zhang, Haowei and Zhang, Mingchuan and Li, YK and Wu, Yang and others},
  journal={arXiv preprint arXiv:2402.03300},
  year={2024}
}

@misc{zhang2026cm2reinforcementlearningchecklist,
      title={CM2: Reinforcement Learning with Checklist Rewards for Multi-Turn and Multi-Step Agentic Tool Use}, 
      author={Zhen Zhang and Kaiqiang Song and Xun Wang and Yebowen Hu and Weixiang Yan and Chenyang Zhao and Henry Peng Zou and Haoyun Deng and Sathish Reddy Indurthi and Shujian Liu and Simin Ma and Xiaoyang Wang and Xin Eric Wang and Song Wang},
      year={2026},
      eprint={2602.12268},
      archivePrefix={arXiv},
      primaryClass={cs.AI},
      url={https://arxiv.org/abs/2602.12268}, 
}

@article{suri2025structured,
  title={Structured Uncertainty guided Clarification for LLM Agents},
  author={Suri, Manan and Mathur, Puneet and Lipka, Nedim and Dernoncourt, Franck and Rossi, Ryan A and Manocha, Dinesh},
  journal={arXiv preprint arXiv:2511.08798},
  year={2025}
}

@article{kadavath2022language,
    title={Language Models (Mostly) Know What They Know},
    author={Kadavath, Saurav and Conerly, Tom and Askell, Amanda and Henighan, Tom and Drain, Dawn and Perez, Ethan and Schiefer, Nicholas and Hatfield-
  Dodds, Zac and DasSarma, Nova and Tran-Johnson, Eli and others},
    journal={arXiv preprint arXiv:2207.05221},
    year={2022}
  }

@article{lin2022teaching,
title={Teaching Models to Express Their Uncertainty in Words},
author={Lin, Stephanie and Hilton, Jacob and Evans, Owain},
journal={arXiv preprint arXiv:2205.14334},
year={2022}
}

@article{kuhn2023semantic,
  title={Semantic uncertainty: Linguistic invariances for uncertainty estimation in natural language generation},
  author={Kuhn, Lorenz and Gal, Yarin and Farquhar, Sebastian},
  journal={arXiv preprint arXiv:2302.09664},
  year={2023}
}

@inproceedings{manakul2023selfcheckgpt,
  title={Selfcheckgpt: Zero-resource black-box hallucination detection for generative large language models},
  author={Manakul, Potsawee and Liusie, Adian and Gales, Mark},
  booktitle={Proceedings of the 2023 conference on empirical methods in natural language processing},
  pages={9004--9017},
  year={2023}
}

@article{kirchhof2025position,
  title={Position: Uncertainty quantification needs reassessment for large-language model agents},
  author={Kirchhof, Michael and Kasneci, Gjergji and Kasneci, Enkelejda},
  journal={arXiv preprint arXiv:2505.22655},
  year={2025}
}

@article{oh2026uncertainty,
title={Uncertainty Quantification in LLM Agents: Foundations, Emerging Challenges, and Opportunities},
author={Oh, Changdae and Park, Seongheon and Kim, To Eun and Li, Jiatong and Li, Wendi and Yeh, Samuel and Du, Xuefeng and Hassani, Hamed and Bogdan,
Paul and Song, Dawn and Li, Sharon},
journal={arXiv preprint arXiv:2602.05073},
year={2026}
}

@article{duan2025uprop,
  title={Uprop: Investigating the uncertainty propagation of llms in multi-step agentic decision-making},
  author={Duan, Jinhao and Diffenderfer, James and Madireddy, Sandeep and Chen, Tianlong and Kailkhura, Bhavya and Xu, Kaidi},
  journal={arXiv preprint arXiv:2506.17419},
  year={2025}
}

@article{zhang2026agentic,
title={Agentic Uncertainty Quantification},
author={Zhang, Jiaxin and Choubey, Prafulla Kumar and Huang, Kung-Hsiang and Xiong, Caiming and Wu, Chien-Sheng},
journal={arXiv preprint arXiv:2601.15703},
year={2026}
}

@article{zhang2026selaur,
title={{SELAUR}: Self Evolving LLM Agent via Uncertainty-aware Rewards},
author={Zhang, Dengjia and Liu, Xiaoou and Cheng, Lu and Wang, Yaqing and Murray, Kenton and Wei, Hua},
journal={arXiv preprint arXiv:2602.21158},
year={2026}
}

@inproceedings{lu2025toolsandbox,
  title={Toolsandbox: A stateful, conversational, interactive evaluation benchmark for llm tool use capabilities},
  author={Lu, Jiarui and Holleis, Thomas and Zhang, Yizhe and Aumayer, Bernhard and Nan, Feng and Bai, Haoping and Ma, Shuang and Ma, Shen and Li, Mengyu and Yin, Guoli and others},
  booktitle={Findings of the Association for Computational Linguistics: NAACL 2025},
  pages={1160--1183},
  year={2025}
}

@inproceedings{patil2025bfcl,
title={The Berkeley Function Calling Leaderboard ({BFCL}): From Tool Use to Agentic Evaluation of Large Language Models},
author={Patil, Shishir G. and Mao, Huanzhi and Yan, Fanjia and Ji, Charlie Cheng-Jie and Suresh, Vishnu and Stoica, Ion and Gonzalez, Joseph E.},
booktitle={Proceedings of the 42nd International Conference on Machine Learning},
pages={48371--48392},
year={2025},
series={Proceedings of Machine Learning Research},
volume={267}
}

@inproceedings{wang2025learning,
title={Learning to Ask: When {LLM} Agents Meet Unclear Instruction},
author={Wang, Wenxuan and Juluan, Shi and Ling, Zixuan and Chan, Yuk-Kit and Wang, Chaozheng and Lee, Cheryl and Yuan, Youliang and Huang, Jen-tse and
Jiao, Wenxiang and Lyu, Michael R.},
booktitle={Proceedings of the 2025 Conference on Empirical Methods in Natural Language Processing},
pages={21773--21784},
year={2025},
publisher={Association for Computational Linguistics},
doi={10.18653/v1/2025.emnlp-main.1104}
}

@article{wu2026tocall,
title={To Call or Not to Call: A Framework to Assess and Optimize LLM Tool Calling},
author={Wu, Qinyuan and Das, Soumi and Amani, Mahsa and Nag, Arijit and Lee, Seungeon and Gummadi, Krishna P. and Ravichander, Abhilasha and Zafar,
Muhammad Bilal},
journal={arXiv preprint arXiv:2605.00737},
year={2026}
}

@article{yang2025qwen3,
  title={Qwen3 technical report},
  author={Yang, An and Li, Anfeng and Yang, Baosong and Zhang, Beichen and Hui, Binyuan and Zheng, Bo and Yu, Bowen and Gao, Chang and Huang, Chengen and Lv, Chenxu and others},
  journal={arXiv preprint arXiv:2505.09388},
  year={2025}
}

@misc{minimax,
  title        = {MiniMax-M2.5},
  author       = {MiniMax},
  year         = {2026},
  url          = {https://huggingface.co/MiniMaxAI/MiniMax-M2.5},
  note         = {Hugging Face Repository},
  organization = {MiniMax}
}

@misc{nemotron,
  title        = {Nemotron-Post-Training-Dataset-v1},
  author       = {Dhruv Nathawani and Igor Gitman and Somshubra Majumdar and Evelina Bakhturina and Ameya Sunil Mahabaleshwarkar and Jian Zhang and Jane Polak Scowcroft},
  year         = {2025},
  url          = {https://huggingface.co/datasets/nvidia/Nemotron-Post-Training-Dataset-v1},
  note         = {Hugging Face Repository},
  organization = {Nvidia}
}

@article{hurst2024gpt,
  title={Gpt-4o system card},
  author={Hurst, Aaron and Lerer, Adam and Goucher, Adam P and Perelman, Adam and Ramesh, Aditya and Clark, Aidan and Ostrow, AJ and Welihinda, Akila and Hayes, Alan and Radford, Alec and others},
  journal={arXiv preprint arXiv:2410.21276},
  year={2024}
}

@article{sun2026when2tool,
  title={LLM Agents Already Know When to Call Tools--Even Without Reasoning},
  author={Sun, Chung-En and Liu, Linbo and Yan, Ge and Wang, Zimo and Weng, Tsui-Wei},
  journal={arXiv preprint arXiv:2605.09252},
  year={2026}
}

@article{modecrua2026multiturn,
  title={Multi-Turn Reinforcement Learning for Tool-Calling Agents with Iterative Reward Calibration},
  author={Modecrua, Wachiravit and Kaewtawee, Krittanon and Pachtrachai, Krittin and Kraisingkorn, Touchapon},
  journal={arXiv preprint arXiv:2604.02869},
  year={2026}
}

@article{zhong2026rcgrpo,
  title={RC-GRPO: Reward-Conditioned Group Relative Policy Optimization for Multi-Turn Tool Calling Agents},
  author={Zhong, Haitian and Zhai, Jixiu and Song, Lei and Bian, Jiang and Liu, Qiang and Tan, Tieniu},
  journal={arXiv preprint arXiv:2602.03025},
  year={2026}
}

@article{healy2026toolselection,
  title={Internal Representations as Indicators of Hallucinations in Agent Tool Selection},
  author={Healy, Kait and Srinivasan, Bharathi and Madathil, Visakh and Wu, Jing},
  journal={arXiv preprint arXiv:2601.05214},
  year={2026}
}

@article{lin2025agenthallusurvey,
  title={LLM-based Agents Suffer from Hallucinations: A Survey of Taxonomy, Methods, and Directions},
  author={Lin, Xixun and Ning, Yucheng and Zhang, Jingwen and Dong, Yan and Liu, Yilong and Wu, Yongxuan and Qi, Xiaohua and Sun, Nan and Shang, Yanmin and Cao, Pengfei and others},
  journal={arXiv preprint arXiv:2509.18970},
  year={2025}
}

@article{su2025autonomyrisk,
  title={A Survey on Autonomy-Induced Security Risks in Large Model-Based Agents},
  author={Su, Hang and Luo, Jun and Liu, Chang and Yang, Xiao and Zhang, Yichi and Dong, Yinpeng and Zhu, Jun},
  journal={arXiv preprint arXiv:2506.23844},
  year={2025}
}

@article{zhang2025agentsafetybench,
  title={Agent-SafetyBench: Evaluating the Safety of LLM Agents},
  author={Zhang, Zhexin and Cui, Shiyao and Lu, Yida and Zhou, Jingzhuo and Yang, Junxiao and Wang, Hongning and Huang, Minlie},
  journal={arXiv preprint arXiv:2412.14470},
  year={2025}
}

@misc{stoisser2025agentsknowdontknow,
      title={Towards Agents That Know When They Don't Know: Uncertainty as a Control Signal for Structured Reasoning}, 
      author={Josefa Lia Stoisser and Marc Boubnovski Martell and Lawrence Phillips and Gianluca Mazzoni and Lea Mørch Harder and Philip Torr and Jesper Ferkinghoff-Borg and Kaspar Martens and Julien Fauqueur},
      year={2025},
      eprint={2509.02401},
      archivePrefix={arXiv},
      primaryClass={cs.AI},
      url={https://arxiv.org/abs/2509.02401}, 
}

@misc{bercovich2025llamanemotronefficientreasoningmodels,
      title={Llama-Nemotron: Efficient Reasoning Models}, 
      author={Akhiad Bercovich and Itay Levy and Izik Golan and Mohammad Dabbah and Ran El-Yaniv and Omri Puny and Ido Galil and Zach Moshe and Tomer Ronen and Najeeb Nabwani and Ido Shahaf and Oren Tropp and Ehud Karpas and Ran Zilberstein and Jiaqi Zeng and Soumye Singhal and Alexander Bukharin and Yian Zhang and Tugrul Konuk and Gerald Shen and Ameya Sunil Mahabaleshwarkar and Bilal Kartal and Yoshi Suhara and Olivier Delalleau and Zijia Chen and Zhilin Wang and David Mosallanezhad and Adi Renduchintala and Haifeng Qian and Dima Rekesh and Fei Jia and Somshubra Majumdar and Vahid Noroozi and Wasi Uddin Ahmad and Sean Narenthiran and Aleksander Ficek and Mehrzad Samadi and Jocelyn Huang and Siddhartha Jain and Igor Gitman and Ivan Moshkov and Wei Du and Shubham Toshniwal and George Armstrong and Branislav Kisacanin and Matvei Novikov and Daria Gitman and Evelina Bakhturina and Prasoon Varshney and Makesh Narsimhan and Jane Polak Scowcroft and John Kamalu and Dan Su and Kezhi Kong and Markus Kliegl and Rabeeh Karimi Mahabadi and Ying Lin and Sanjeev Satheesh and Jupinder Parmar and Pritam Gundecha and Brandon Norick and Joseph Jennings and Shrimai Prabhumoye and Syeda Nahida Akter and Mostofa Patwary and Abhinav Khattar and Deepak Narayanan and Roger Waleffe and Jimmy Zhang and Bor-Yiing Su and Guyue Huang and Terry Kong and Parth Chadha and Sahil Jain and Christine Harvey and Elad Segal and Jining Huang and Sergey Kashirsky and Robert McQueen and Izzy Putterman and George Lam and Arun Venkatesan and Sherry Wu and Vinh Nguyen and Manoj Kilaru and Andrew Wang and Anna Warno and Abhilash Somasamudramath and Sandip Bhaskar and Maka Dong and Nave Assaf and Shahar Mor and Omer Ullman Argov and Scot Junkin and Oleksandr Romanenko and Pedro Larroy and Monika Katariya and Marco Rovinelli and Viji Balas and Nicholas Edelman and Anahita Bhiwandiwalla and Muthu Subramaniam and Smita Ithape and Karthik Ramamoorthy and Yuting Wu and Suguna Varshini Velury and Omri Almog and Joyjit Daw and Denys Fridman and Erick Galinkin and Michael Evans and Shaona Ghosh and Katherine Luna and Leon Derczynski and Nikki Pope and Eileen Long and Seth Schneider and Guillermo Siman and Tomasz Grzegorzek and Pablo Ribalta and Monika Katariya and Chris Alexiuk and Joey Conway and Trisha Saar and Ann Guan and Krzysztof Pawelec and Shyamala Prayaga and Oleksii Kuchaiev and Boris Ginsburg and Oluwatobi Olabiyi and Kari Briski and Jonathan Cohen and Bryan Catanzaro and Jonah Alben and Yonatan Geifman and Eric Chung},
      year={2025},
      eprint={2505.00949},
      archivePrefix={arXiv},
      primaryClass={cs.CL},
      url={https://arxiv.org/abs/2505.00949}, 
}

@article{lymperopoulos2025tools,
  title={Tools in the loop: Quantifying uncertainty of llm question answering systems that use tools},
  author={Lymperopoulos, Panagiotis and Sarathy, Vasanth},
  journal={arXiv preprint arXiv:2505.16113},
  year={2025}
}

@article{lu2026homeguard,
  title={HomeGuard: VLM-based Embodied Safeguard for Identifying Contextual Risk in Household Task},
  author={Lu, Xiaoya and Zhou, Yijin and Chen, Zeren and Wang, Ruocheng and Sima, Bingrui and Zhou, Enshen and Sheng, Lu and Liu, Dongrui and Shao, Jing},
  journal={arXiv preprint arXiv:2603.14367},
  year={2026}
}

@article{zhou2026infa,
  title={INFA-Guard: Mitigating Malicious Propagation via Infection-Aware Safeguarding in LLM-Based Multi-Agent Systems},
  author={Zhou, Yijin and Lu, Xiaoya and Liu, Dongrui and Yan, Junchi and Shao, Jing},
  journal={arXiv preprint arXiv:2601.14667},
  year={2026}
}

@inproceedings{lu2026bench,
  title={Is-bench: Evaluating interactive safety of vlm-driven embodied agents in daily household tasks},
  author={Lu, Xiaoya and Chen, Zeren and Hu, Xuhao and Zhou, Yijin and Zhang, Weichen and Liu, Dongrui and Sheng, Lu and Shao, Jing},
  booktitle={Proceedings of the AAAI Conference on Artificial Intelligence},
  volume={40},
  number={42},
  pages={35680--35688},
  year={2026}
}

@article{zhou2024out,
  title={How Out-of-Distribution Detection Learning Theory Enhances Transformer: Learnability and Reliability},
  author={Zhou, Yijin and Ge, Yutang and Xie, Wenyuan and Zeng, Linqian and Dong, Xiaowen and Wang, Yuguang},
  journal={arXiv preprint arXiv:2406.12915},
  year={2024}
}

@inproceedings{chen2026geometrically,
  title={Geometrically-constrained agent for spatial reasoning},
  author={Chen, Zeren and Lu, Xiaoya and Zheng, Zhijie and Li, Pengrui and He, Lehan and Zhou, Yijin and Shao, Jing and Zhuang, Bohan and Sheng, Lu},
  booktitle={Proceedings of the IEEE/CVF Conference on Computer Vision and Pattern Recognition},
  pages={38689--38699},
  year={2026}
}
}
% \clearpage
\appendix

\section{Details of Benchmarks and Baselines}
\label{app: Baselines}
\subsection{Benchmarks}
We evaluate three benchmarks for \sys~ framework:
\begin{itemize}
    \item When2Call \citep{ross2025when2call}: a natural benchmark for next-action decision training and testing, as it explicitly evaluates whether an agent should answer directly, call a tool, ask a follow-up question, or admit that the provided tools cannot answer the request.
    \item ToolSandbox \citep{lu2025toolsandbox}: evaluates stateful, conversational, and interactive tool use, with scenarios such as canonicalization, tool scrambled, and insufficient information that require the agent to decide when and how to execute tools versus ask or wait for more context.
    \item BFCL-V4 \citep{patil2025bfcl}: evaluates function-calling reliability across single, multiple, parallel, multi-turn, and relevance-detection settings, where irrelevant or unsuitable function sets directly test whether the model can abstain from unsupported tool calls.
\end{itemize}

\subsection{Baselines}
We compare against two training-free uncertainty-aware agent baselines:
\begin{itemize}
    \item \textbf{AUQ} \citep{zhang2026agentic}: uses verbalized confidence and explanations to control long-horizon agent execution, propagating uncertainty through memory and invoking reflection only when confidence is low.
    \item \textbf{SAGE} \citep{suri2025structured}: a structured uncertainty-guided clarification framework that models uncertainty over tool calls and argument domains, then uses EVPI-based scoring to decide whether to ask a targeted clarification question or execute the best tool call.
\end{itemize}

\section{Detailed Setups}
\label{app: Prompts and Settings}
\subsection{Prompts}
Prompts for the LLM labeler can be found in Box~\ref{box: label}, and prompts for the LLM judger in the assigning tool-calling decision reward can be found in Box~\ref{box: judger}. The judger is aiming to judge the decision category in the response of agents, rather than obtain the reward $R_{\text{UQ}}$ in one step.

\subsection{Training Settings}
For turn-level training based on Qwen3-4B-Thinking, we train 1 epoch GRPO. We sample 4 rollouts for each question as the group size. The learning rate is set to $3\times10^{-5}$, and the KL divergence loss coefficient is set to 0.001. 
For trajectory-level training based on Qwen3-8B-Base, we follow the training settings in \citet{zhang2026cm2reinforcementlearningchecklist} for unified post-training, including cold-start SFT and RL. 
All training data is from the \textsc{nvidia/Nemotron-Post-Training-Dataset-v1} dataset \citet{bercovich2025llamanemotronefficientreasoningmodels, nemotron} and filtered by \citet{zhang2026cm2reinforcementlearningchecklist}. \textsc{nvidia/Nemotron-Post-Training-Dataset-v1} contains a large number of 310k synthetic tool-use dialogues across diverse domain.
For SFT, we use the cold-start training set in \citet{zhang2026cm2reinforcementlearningchecklist}, which has 8k trajectories. We apply SFT as a cold start under a learning rate $3\times10^{-6}$ with a warmup ratio of 0.1. For RL, the corresponding training set in \citet{zhang2026cm2reinforcementlearningchecklist} is further annotated by \sys~ and filtered for turn-level decision distribution balance. Finally, we get 4k trajectories from the original 8k data. Then we optimize from the cold-start SFT checkpoint using GRPO based on VeRL. The mini-batch size is set to 128, the KL divergence loss coefficient to 0.001, and the learning rate is $3\times10^{-6}$ for GRPO. The group size is 16 for one trajectory. We train 400 GPU hours for GRPO. All experiments are conducted on $8 \times$ NVIDIA H200 GPUs using DeepSpeed ZeRO-3 optimization.

\section{Additional Exprimental Results}
\label{app: results}

Detailed experimental results on BFCL-V4 can be found in Tables~\ref{tab:bfcl_non-live}--\ref{tab:bfcl_agentic}, where BFCL-V4 is decomposed into four categories: Non-Live (Table~\ref{tab:bfcl_non-live}), Live (Table~\ref{tab:bfcl_live}), Multi-Turn (Table~\ref{tab:bfcl_multi-turn}), and Agentic (Table~\ref{tab:bfcl_agentic}). Ablation study results on When2Call are presented in the Table~\ref{tab:ablation_results_when2call}. For trajectory-level post-training from Qwen3-8B-Base, ablation results on BFCL-V4 are further broken down into Non-Live (Table~\ref{tab:bfcl_ablation_non-live}), Live (Table~\ref{tab:bfcl_ablation_live}), Multi-Turn (Table~\ref{tab:bfcl_ablation_multi-turn}), and Agentic (Table~\ref{tab:bfcl_ablation_agentic}) categories. Ablation results on ToolSandBox are provided in Table~\ref{tab:toolsandbox_ablation}.

\subsection{Results on BFCL-V4}

Tables~\ref{tab:bfcl_non-live}--\ref{tab:bfcl_agentic} report detailed results on BFCL-V4 across Non-Live, Live, Multi-Turn, and Agentic categories. Overall, \sys~ consistently improves tool-calling performance under both turn-level and trajectory-level training settings, with particularly large gains on complex multi-turn and agentic scenarios.

\paragraph{Non-Live and Live tool calling.}
On relatively direct tool-calling categories, such as Non-Live and Live, the benefit of trajectory-level optimization is already evident. As shown in Tables~\ref{tab:bfcl_non-live} and~\ref{tab:bfcl_live}, Traj.-level \sys~ improves over Traj.-level CM2 from 80.12\% to 82.27\% on Non-Live, and from 74.58\% to 75.57\% on Live. The gains are moderate but consistent, suggesting that these categories can be effectively optimized with trajectory-level feedback.

The ablation results further show that $c(s)$ is less critical for these relatively simple single-turn settings. For example, in Tables~\ref{tab:bfcl_ablation_non-live} and \ref{tab:bfcl_ablation_live}, +\sys~ w/o $c(s)$ already reaches 78.90\% on Non-Live and 74.17\% on Live, which are close to the SFT or CM2 baselines. On some subcategories, such as Live Parallel Multiple, +\sys~ w/o $c(s)$ even matches the full \sys~ score of 54.17\%. These results suggest that for direct tool-calling tasks where the decision boundary is relatively clear, trajectory-level answer and format rewards can already provide useful learning signals, even without an explicit classification reward.

\paragraph{Multi-turn tool calling.}
The advantage of \sys~ becomes more pronounced in Multi-Turn scenarios, which require maintaining context, resolving missing information, and making correct tool-use decisions across turns. As shown in Table~\ref{tab:bfcl_multi-turn}, Traj.-level \sys~ improves over Traj.-level CM2 from 22.50\% to 29.88\%, yielding a substantial gain of 7.38\% points. The improvement is consistent across all subcategories: Multi Turn Base improves from 34.00\% to 40.50\%, Missing Function from 21.00\% to 26.50\%, Missing Parameter from 18.00\% to 28.00\%, and Long Context from 17.00\% to 24.50\%.

These gains indicate that \sys~ is particularly effective in higher-order tool-calling settings where the model must reason over dialogue history and incomplete user intents. Compared with simpler Non-Live and Live categories, Multi-Turn tasks rely more heavily on correctly identifying whether, when, and how to invoke tools. Therefore, the uncertainty repulsive reward $cR_{\text{cls}}$ becomes more important: without it, the model may still learn from successful trajectories, but it lacks a direct signal for distinguishing tool-use versus non-tool-use decisions across turns.

\paragraph{Agentic tool calling.}
Large trajectory-level gains also appear in the Agentic category, which includes Web Search and Memory tasks. As shown in Table~\ref{tab:bfcl_agentic}, Traj.-level \sys~ significantly improves over Traj.-level CM2 from 21.97\% to 28.79\% on the overall Agentic score, corresponding to a 6.82\% gain. The improvement is especially clear on Memory, where \sys~ improves from 21.94\% to 30.57\%. Within Memory, Key Value Store increases from 9.68\% to 21.00\%, Vector Store from 14.19\% to 24.55\%, and Rec Sum from 41.94\% to 46.16\%.

These results highlight the importance of the proposed reward design for high-level agentic tool use. Agentic tasks require not only producing correct tool arguments, but also deciding when external search, memory retrieval, or memory update is necessary. In such settings, $cR_{\text{cls}}$ provides an essential signal for tool-use decision making. Compared with Non-Live and Live categories, where +\sys~ w/o $cR_{\text{cls}}$ can already obtain reasonable performance, Agentic tasks benefit much more from the full reward. This suggests that explicit classification supervision is critical for complex agent behaviors involving planning, memory, and external information access.

\subsection{Ablation Study on When2Call}

Table~\ref{tab:ablation_results_when2call} presents the ablation study on the When2Call test set. Since When2Call directly evaluates whether the model calls tools at the right time, it provides a focused benchmark for analyzing the contribution of different reward components, especially $c(s)$.

\paragraph{Effect of reward components under turn-level training.}
Starting from Qwen3-4B-Thinking, turn-level \sys~ achieves the best overall Acc Norm of 80.83\% and F1 of 82.84\%, outperforming the base checkpoint by 11.47 points in Acc Norm and 7.07\% points in F1. Removing different reward components leads to different failure modes. Without $R_{\text{ans}}$, the model still obtains a strong Acc Norm of 78.97\%, suggesting that classification and format rewards can already guide the model toward better tool-use decisions. However, without $c(s)$, Acc Norm drops to 72.46\% and FDAR increases to 24.76\%, indicating that the model becomes worse at deciding when not to call tools. Without $R_{\text{fmt}}$, the model obtains 75.90\% Acc Norm, but Tool Hall remains relatively high at 24.03\%.

These results show that each reward component contributes differently: $R_{\text{ans}}$ improves final task correctness, $R_{\text{fmt}}$ stabilizes valid tool-call formatting, and $c(s)$ is especially important for suppressing incorrect tool-use decisions. The sharp FDAR increase after removing $c(s)$ confirms that classification reward plays a key role in calibrating when tools should or should not be invoked.

\paragraph{Trajectory-level ablation from Qwen3-8B-Base.}
The importance of $c(s)$ is even clearer in the trajectory-level setting. Starting from Qwen3-8B-Base, SFT only achieves 19.20\% Acc Norm and 27.29 F1\%, showing that supervised fine-tuning alone is insufficient under this evaluation protocol. Adding Traj.-level CM2 improves Acc Norm to 43.75\%, but it also produces a very high Tool Hall score of 75.43\%, suggesting severe over-calling or hallucinated tool use.

In contrast, + Traj.-level \sys~ w/o $c(s)$ only reaches 31.43\% Acc Norm and 26.19\% F1, with FDAR increasing to 27.14\%. This indicates that trajectory-level optimization without classification reward provides limited guidance for deciding whether tools should be used. The full + Traj.-level \sys~ substantially improves Acc Norm to 62.32\% and F1 to 60.62\%, while reducing Tool Hall to 49.87\% and FDAR to 8.39\%. Compared with the w/o $c(s)$ variant, the full reward improves Acc Norm by 30.89 points and F1 by 34.43 points.

These results demonstrate that $c(s)$ is crucial for When2Call-style decision making. While w/o $c(s)$ can perform reasonably well on simpler BFCL categories such as Non-Live and Live, it is insufficient for benchmarks that explicitly require accurate tool-use timing. The full \sys~ reward is therefore necessary for learning robust tool-use policies that balance calling tools when needed and avoiding unnecessary or hallucinated tool calls.

\section{Case Study}
\label{app: case}
\subsection{Case Study 1: Correct Tool Parameter Usage}
\begin{figure*}
    \centering
    \includegraphics[width=\linewidth]{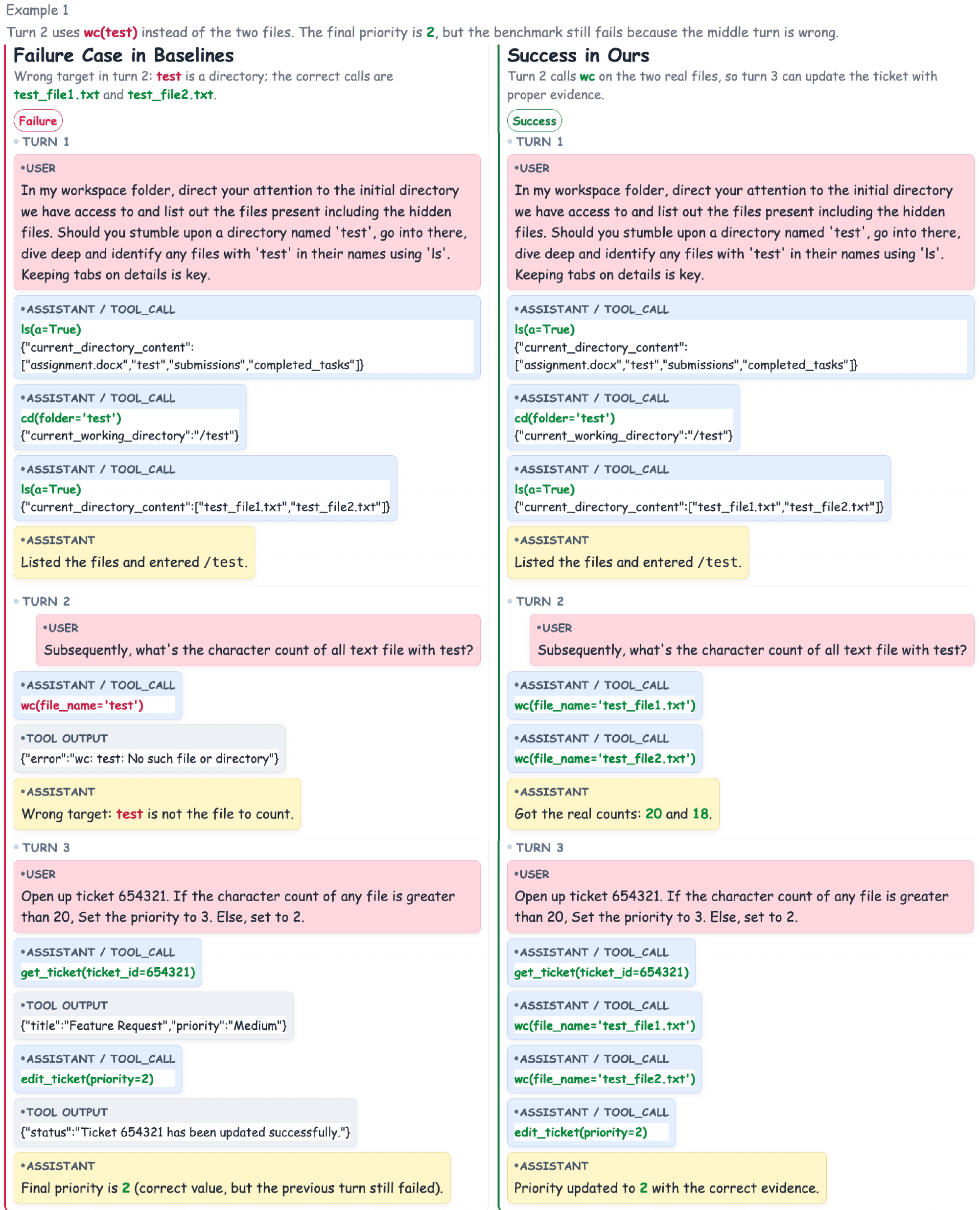}
    \caption{Example of tool-calling parameter failure in baselines and success in our method.}
    \label{fig:6-ex1}
\end{figure*}

As shown in Fig.~\ref{fig:6-ex1}, a failure mode is one in which the agent reaches a superficially correct final decision while still failing the task at the trajectory level. After correctly entering the \texttt{test} directory and identifying \texttt{test\_file1.txt} and \texttt{test\_file2.txt}, the baseline incorrectly applies \texttt{wc} to the directory name \texttt{test} rather than to the two discovered files. As a result, the required intermediate evidence, namely the character counts of the relevant text files, is never obtained. Nevertheless, the model proceeds to update the ticket priority to 2, which happens to coincide with the correct final value because the true counts (20 and 18) are both not greater than 20. In contrast, \sys~ executes the intended file-level counting operations and therefore reaches the same final priority assignment through a valid evidential chain.

\subsection{Case Study 2: Timely Tool Invocation}

\begin{figure*}
    \centering
    \includegraphics[width=0.75\linewidth]{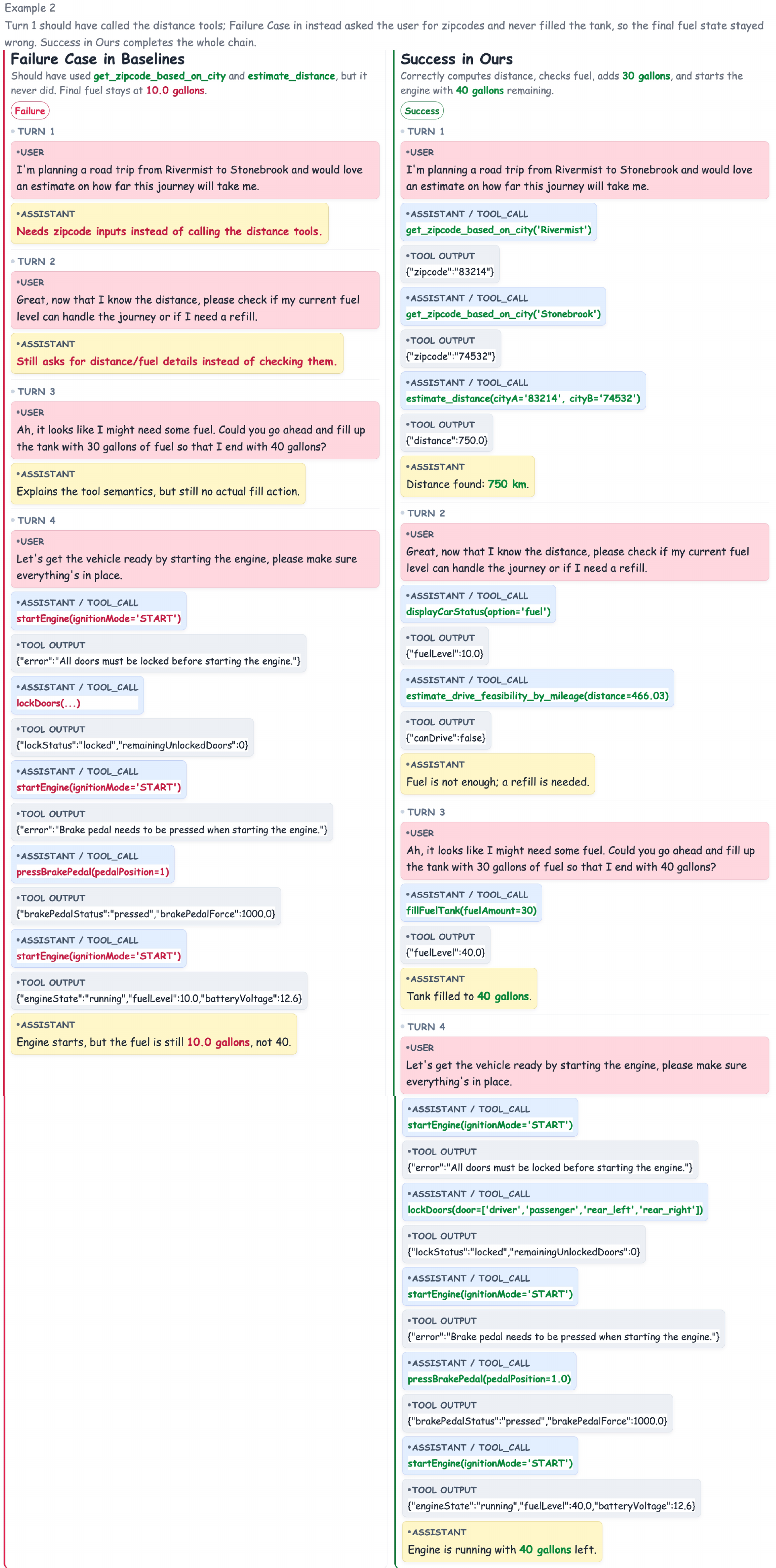}
    \caption{Example of tool-calling decision failure of missed tool invocation and downstream task collapse, and success in our method.}
    \label{fig:6-ex2}
\end{figure*}

The second case study (Fig.~\ref{fig:6-ex2}) demonstrates a complementary failure pattern in the baseline: the agent abstains from calling tools when tool use is necessary to advance the task. A missing tool call in an early turn can destabilize the entire downstream interaction and produce a globally incorrect final state. In the first turn, the baseline requests zip code information from the user instead of invoking the available city-to-zipcode and distance estimation tools. This omission prevents the model from grounding the subsequent reasoning about travel feasibility. The error compounds in later turns: the model does not execute the requested fuel-filling action, and by the final turn, the vehicle state still reports 10.0 gallons, directly contradicting the user’s requirement to add 30 gallons and end with 40 gallons. By comparison, \sys~ correctly performs the tool chain for geographic lookup, distance estimation, fuel feasibility assessment, tank filling, and engine start, uses the right tool in key turns, and completes the task.

\subsection{Case Study 3: Deferring Tool Use under Missing Information}

\begin{figure*}
    \centering
    \includegraphics[width=0.9\linewidth]{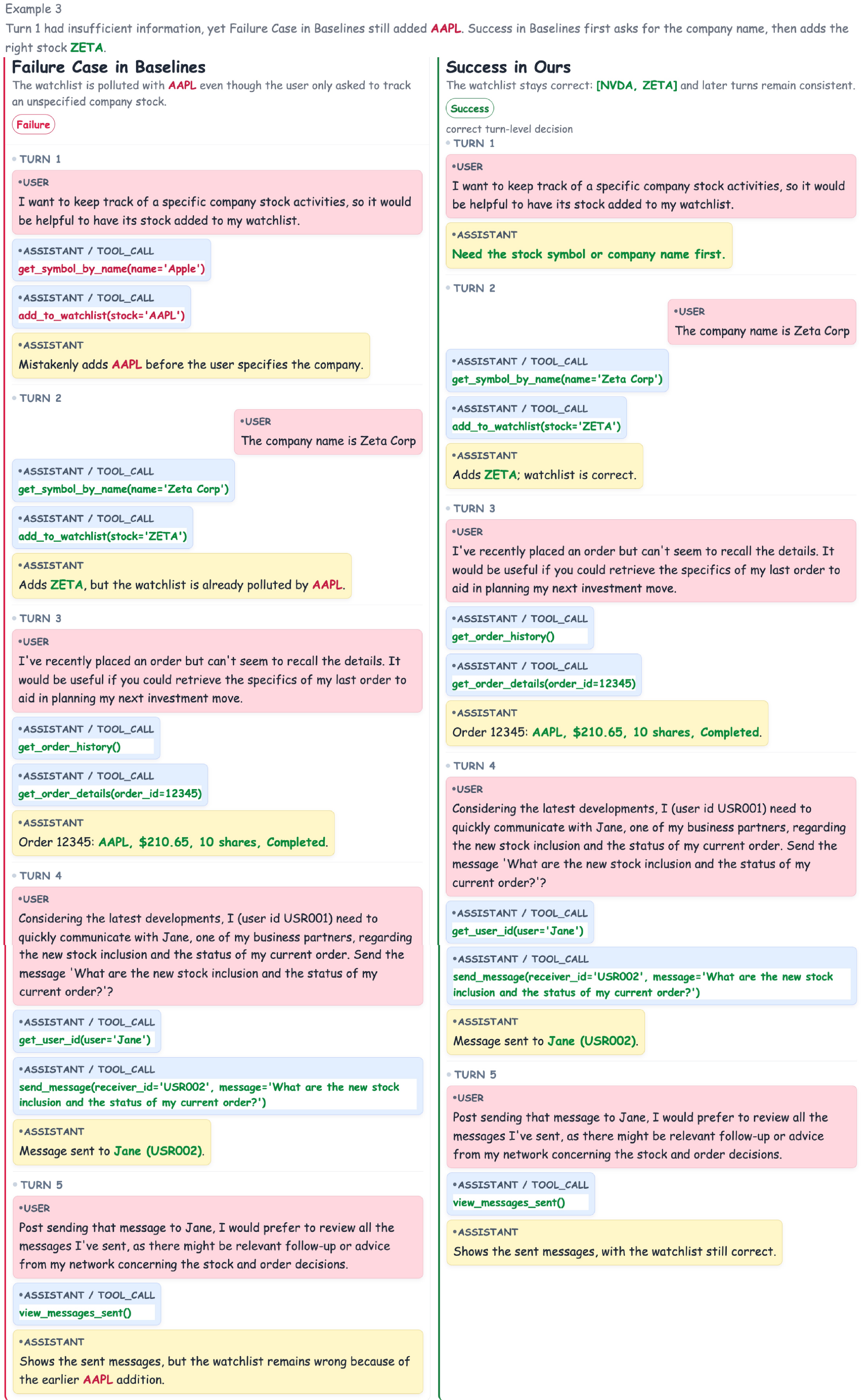}
    \caption{Example of tool-calling decision failure of unwarranted tool use under information insufficiency, and success in our method.}
    \label{fig:6-ex3}
\end{figure*}

Example in Fig.~\ref{fig:6-ex3} provides a clearer example of inappropriate tool invocation under incomplete user specification. In the first turn, the user expresses the intent to add a company’s stock to the watchlist but does not provide the company name. The benchmark’s reference trajectory, therefore, leaves this turn empty, indicating that the correct behavior is to defer action and elicit the missing information. Instead, the baseline hallucinates the company as Apple, invokes \texttt{get\_symbol\_by\_name}, and adds \texttt{AAPL} to the watchlist. Although the model later adds \texttt{ZETA} after the user explicitly specifies ``Zeta Corp'', the watchlist has already been irreversibly contaminated with an incorrect extra entry. The agent trained by our method \sys, in contrast, handles the underspecified first turn conservatively by requesting the missing company name and only performs tool calls after the parameter becomes available.

\onecolumn
\begin{promptbox}[label=box: label]{Prompt for Labeling Json Annotation}
\#\# You are annotating multi-turn tool-use trajectories for when2call training, namely supervision for deciding when to answer directly, invoke tools, request missing information, or abstain.
\newline
\newline
\#\# Task: 
\newline
1. Read the full trajectory, the available tool specifications, and the checklist metadata.
\newline
2. Identify decision-critical turns at which a when2call-style supervision signal is both informative and well supported by the observed trajectory.
\newline
3. For each selected turn, assign exactly one ground-truth action from the following action space:
\newline
   - `direct\_answer`: the request can be correctly addressed from the existing conversational context or general knowledge, without additional user input and without tool use.
\newline
   - `tool\_call`: the appropriate next step is to invoke one or more tools, and the necessary arguments are already available.
\newline
   - `request\_for\_info`: the assistant lacks essential user-provided information, and clarification is the correct immediate next action.
\newline
   - `cannot\_answer`: the assistant should refuse, abstain, or acknowledge incapability or lack of access, such that further clarification would not resolve the problem.
\newline
4. Prefer substantive reasoning over shallow lexical heuristics.
\newline
5. Maintain broad coverage over the four action categories; do not overproduce `tool\_call` annotations merely because the data originate from a tool-use setting.
\newline
\newline
\#\# Annotation principles:
\newline
- Prioritize turns that are genuinely decision-critical, including missing arguments, unsupported or hallucinated tool usage, inappropriate tool invocation, clarification after tool failure, directly answerable requests, and authentic inability cases.
\newline
- When multiple `turn\_idx` values are defensible within the same trajectory, prefer the one whose correct action would most improve action-space coverage under the provided balance guidance.
\newline
- Avoid producing many redundant annotations of the same class from a single trajectory, especially repeated `request\_for\_info` cases.
\newline
- Do not default to `request\_for\_info` as a conservative fallback. If the assistant already has sufficient information to answer, choose `direct\_answer`. If the assistant already has sufficient information to act, choose `tool\_call`. If no realistic clarification would resolve the issue, choose `cannot\_answer`.
\newline
- Use the entire trajectory, including earlier tool outputs and prior assistant behavior. Later turns often support `direct\_answer` or `tool\_call` because the required information has already been established upstream.
\newline
- The `target\_answer` must remain compatible with the when2call reward interface:
\newline
  (1) For `tool\_call`, return one or more concrete tool-call XML blocks in the form `<tool\_call>\{"name": "...", "arguments": \{...\}\}</tool\_call>`.
\newline
  (2) For `request\_for\_info`, return the realized clarification question.
\newline
  (3) For `direct\_answer`, return the realized answer text.
\newline
  (4) For `cannot\_answer`, return the realized refusal or inability statement.
\newline
\newline

\#\# Illustrative cases:
\newline
- `direct\_answer`: the user requests an explanation, rewrite, or straightforward factual transformation that can already be completed without tools.
\newline
- `tool\_call`: the user has supplied all required slots, and the correct next action is immediate tool execution.
\newline
- `request\_for\_info`: a decisive slot required for answering or tool execution is missing from the user.
\newline
- `cannot\_answer`: the available toolset cannot solve the request, or the assistant lacks the necessary authority, capability, or access, and clarification would not remedy the limitation.

\#\# Return strict JSON:
\newline
\{\newline
  "turn\_idx": Current `turn\_idx` in data,\newline
  "gt\_action": Action for this turn in `direct\_answer`/`tool\_call`/`request\_for\_info`/`cannot\_answer`,\newline
  "rationale": Your explanation for `gt\_action`,\newline
  "target\_answer": Real response in data\newline
\}
\newline
\newline
\#\# Here are two illustrative annotation examples:
\newline
- Example 1:
\newline
\{\newline
  "turn\_idx": 2,\newline
  "gt\_action": "request\_for\_info",\newline
  "rationale": "Missing province ID required to continue the request correctly.",\newline
  "target\_answer": "Could you please provide me with the specific province ID for Bangkok?"\newline
\}
\newline
- Example 2:
\newline
\{\newline
  "turn\_idx": 0,\newline
  "gt\_action": "tool\_call",\newline
  "rationale": "All required parameters are available. The assistant should call the tool immediately.",\newline
  "target\_answer": "<tool\_call>{"name": "a", "arguments": {"account\_id": "123"}}</tool\_call>"\newline
    \}
\newline \newline

\#\# Dataset balance guidance: \newline
- Current accepted annotation counts: `direct\_answer`=\{DIRECT\_ANSWER\_COUNT\}, `tool\_call`=\{TOOL\_CALL\_COUNT\}, `request\_for\_info`=\{REQUEST\_FOR\_INFO\_COUNT\}, `cannot\_answer`=\{CANNOT\_ANSWER\_COUNT\}\newline
- Currently underrepresented action categories: \{UNDERREPRESENTED\_ACTIONS\}\newline
- Currently overrepresented action categories: \{OVERREPRESENTED\_ACTIONS\}\newline
- If multiple `turn\_idx` values within the same trajectory are equally well justified, prefer the one whose correct `gt\_action` belongs to an underrepresented category.\newline
- Prefer valid non-`tool\_call` annotations when `tool\_call` is already overrepresented.\newline
- Favor a small number of high-value annotations per trajectory rather than many redundant ones.\newline
- Do not force an annotation whose action is not supported by the trajectory evidence.
\newline\newline  

\#\# Trajectory payload:
\{\newline
  "tools": \{TOOLS\_JSON\},\newline
  "turn\_summaries": \{TURN\_SUMMARIES\_JSON\},\newline
  "checklists": \{CHECKLISTS\_JSON\}\newline
\}\newline
\end{promptbox}

\begin{promptbox}[label=box: judger]{Prompt for LLM Judger in \sys}
\#\# You are a structured parser for assistant turns in multi-turn tool-use trajectories.
\newline  \newline  
\#\# Given the history of completed assistant turns within a multi-turn trajectory:\newline  
1. Judge the assistant's next action from the current observed turn.\newline  
2. Map the action to exactly one normalized action category:\newline  
   - `direct\_answer`: the assistant provides a substantive answer directly, without invoking a tool and without requesting additional user information.\newline  
   - `tool\_call`: the assistant initiates one or more tool calls as the immediate next action.\newline  
   - `request\_for\_info`: the assistant asks the user for missing information that is necessary before answering or executing a tool call.\newline  
   - `cannot\_answer`: the assistant indicates inability, refusal, lack of capability, or lack of access such that further clarification would not resolve the issue.\newline  
3. Extract the realized response content in a when2call-compatible representation:\newline  
   - If the action is `tool\_call`, return only the tool-call JSON content, without any surrounding prose.\newline  
   - Otherwise, return the assistant's realized natural-language response exactly as expressed in the turn.\newline  
4. Ground the decision in the realized turn behavior rather than in hypothetical alternatives.
\newline  \newline  
\#\# The annotation of this turn:\newline  
\{JSON\_ANNOTATON\}\newline  

\#\# If the assistant has conducted `gt\_action` in `JSON\_ANNOTATION` in this turn, choose `pred\_action` the same as `gt\_action`.\newline  \newline  

\#\# Return strict JSON. You MUST follow this Output Schema:\newline  
\{\newline  
  "pred\_action": Action for this turn in `direct\_answer`/`tool\_call`/`request\_for\_info`/`cannot\_answer`,\newline  
  "pred\_answer": Response of the assistant in this turn,\newline  
  "rationale": "The turn is a clarification request because a required argument is not yet available. The assistant requests a missing identifier before any tool execution."\newline  
\}\newline  \newline  

\#\# Here is an example:\newline  
\{\newline  
  "pred\_action": "request\_for\_info",\newline  
  "pred\_answer": "Could you provide the account ID?",\newline  
  "rationale": "The turn is a clarification request because a required argument is not yet available. The assistant requests a missing identifier before any tool execution."\newline  
\}\newline  
\end{promptbox}

\twocolumn

% Table 1
\begin{table*}[t]
\centering
\small
\setlength{\tabcolsep}{16pt}
\renewcommand{\arraystretch}{1.15}
\caption{Detailed comparison experiment results of Non-Live category on BFCL-V4.}
\label{tab:bfcl_non-live}
\resizebox{\linewidth}{!}{
\begin{tabular}{l|c|cccc}
\toprule
\textbf{Models} & \textbf{Non-Live} & \textbf{Simple} & \textbf{Multiple} & \textbf{Parallel} & \textbf{Parallel Multiple} \\
\midrule
\multicolumn{6}{l}{{\cellcolor{gray!6}\textit{Open-source Baselines}}}\\
235B-A22B-Instruct & 90.10 & 78.42 & 95.50 & 94.00 & 92.50 \\
30B-A3B-Instruct  & 85.77 & 68.58 & 94.50 & 91.50 & 88.50 \\
4B-Thinking       & 38.02 & 41.58 & 37.00 & 53.50 & 20.00 \\
8B-Thinking       & 87.58 & 72.83 & 96.50 & 92.00 & 89.00 \\
\midrule
\multicolumn{6}{l}{{\cellcolor{gray!6}\textit{From Qwen3-4B-Thinking}}}\\
AUQ                & 77.08 & 66.83 & 81.50 & 80.00 & 80.00 \\
SAGE               & \underline{86.71} & 71.33 & 95.00 & 92.50 & 86.00 \\
Turn-level GRPO    & 86.33 & 73.33 & 94.50 & 91.00 & 86.50 \\
\textbf{Turn-level \sys} & \textbf{89.23} & 77.92 & 96.00 & 92.50 & 90.50 \\ \midrule
\multicolumn{6}{l}{{\cellcolor{gray!6}\textit{From Qwen3-8B-Thinking}}}\\
AUQ                & \underline{84.46} & 70.83 & 90.00 & 92.50 & 84.50 \\
SAGE               & \textbf{88.15} & 74.58 & 95.00 & 93.00 & 90.00 \\
\multicolumn{6}{l}{{\cellcolor{gray!6}\textit{From Qwen3-8B-Base}}}\\
Traj.-level CM2    & 80.12 & 70.50 & 93.00 & 85.00 & 72.00 \\
\textbf{Traj.-level \sys} & 82.27 & 71.08 & 94.50 & 85.00 & 78.50 \\
\bottomrule
\end{tabular}
}
\end{table*}

% Table 2
\begin{table*}[t]
\centering
\small
\setlength{\tabcolsep}{16pt}
\renewcommand{\arraystretch}{1.15}
\caption{Detailed comparison experiment results of Live category on BFCL-V4.}
\label{tab:bfcl_live}
\resizebox{\linewidth}{!}{
\begin{tabular}{l|c|cccc}
\toprule
\textbf{Models} & \textbf{Live} & \textbf{Simple} & \textbf{Multiple} & \textbf{Parallel} & \textbf{Parallel Multiple}  \\
\midrule
\multicolumn{6}{l}{{\cellcolor{gray!6}\textit{Open-source Baselines}}}\\
235B-A22B-Instruct & 83.20 & 87.60 & 82.24 & 75.00 & 83.33  \\
30B-A3B-Instruct  & 77.94 & 83.33 & 76.83 & 68.75 & 75.00  \\
4B-Thinking       & 71.06 & 63.18 & 72.84 & 87.50 & 66.67  \\
8B-Thinking       & 80.53 & 84.50 & 79.68 & 75.00 & 79.17 \\
\midrule
\multicolumn{6}{l}{{\cellcolor{gray!6}\textit{From Qwen3-4B-Thinking}}}\\
AUQ                & 62.40 & 65.89 & 61.06 & 87.50 & 66.67  \\
SAGE               & 80.49 & 83.05 & 79.81 & 84.50 & 80.33  \\
Turn-level GRPO    & \underline{80.61} & 85.66 & 79.58 & 81.25 & 70.83 \\
\textbf{Turn-level \sys} & \textbf{84.09} & 89.53 & 82.62 & 87.50 & 87.50  \\ \midrule
\multicolumn{6}{l}{{\cellcolor{gray!6}\textit{From Qwen3-8B-Thinking}}}\\
AUQ                & 75.42 & 80.56 & 74.26 & 79.25 & 68.83  \\
SAGE               & \textbf{80.83} & 84.88 & 80.06 & 68.75 & 79.17  \\
\multicolumn{6}{l}{{\cellcolor{gray!6}\textit{From Qwen3-8B-Base}}}\\
Traj.-level CM2    & 74.58 & 74.64 & 75.17 & 76.00 & 46.83 \\
\textbf{Traj.-level \sys} & \underline{75.57} & 75.97 & 75.97 & 75.00 & 54.17 \\
\bottomrule
\end{tabular}
}
\end{table*}

% Table 3
\begin{table*}[t]
\centering
\small
\setlength{\tabcolsep}{3pt}
\renewcommand{\arraystretch}{1.15}
\caption{Detailed comparison experiment results of the Multi-turn category on BFCL-V4.}
\label{tab:bfcl_multi-turn}
\resizebox{\linewidth}{!}{
\begin{tabular}{l|c|cccc}
\toprule
\textbf{Models} & \textbf{Multi-Turn} & \textbf{Multi Turn Base} & \textbf{Multi Turn Missing Function} & \textbf{Multi Turn Missing Parameter} & \textbf{Multi Turn Long Context} \\
\midrule
\multicolumn{6}{l}{{\cellcolor{gray!6}\textit{Open-source Baselines}}}\\
235B-A22B-Instruct & 50.13&63.50 & 45.00 & 38.50 & 53.50 \\
30B-A3B-Instruct  & 30.00&43.50 & 10.50 & 25.00 & 41.00 \\
4B-Thinking       & 52.00&63.50 & 58.50 & 35.00 & 51.00 \\
8B-Thinking       & 37.75&46.50 & 38.00 & 36.00 & 30.50 \\
\midrule
\multicolumn{6}{l}{{\cellcolor{gray!6}\textit{From Qwen3-4B-Thinking}}}\\
AUQ                & 23.50&35.00 & 22.00 & 16.00 & 21.00 \\
SAGE               & \underline{41.50} &59.00 & 55.50 & 19.00 & 32.50 \\
Turn-level GRPO    & 22.75&31.00 & 21.50 & 25.00 & 13.50 \\
\textbf{Turn-level \sys} & \textbf{49.63}&60.50 & 58.00 & 45.50 & 34.50 \\ \midrule
\multicolumn{6}{l}{{\cellcolor{gray!6}\textit{From Qwen3-8B-Thinking}}}\\
AUQ                & 10.50&21.00 & 7.50 & 7.50 & 6.00 \\
SAGE               & \textbf{36.38}&47.00 & 44.00 & 29.50 & 25.00 \\
\multicolumn{6}{l}{{\cellcolor{gray!6}\textit{From Qwen3-8B-Base}}}\\
Traj.-level CM2    & 22.50&34.00 & 21.00 & 18.00 & 17.00 \\
\textbf{Traj.-level \sys}  & \underline{29.88} &40.50 & 26.50 & 28.00 & 24.50 \\
\bottomrule
\end{tabular}
}
\end{table*}

% Table 4
\begin{table*}[t]
\centering
\small
\setlength{\tabcolsep}{5pt}
\renewcommand{\arraystretch}{1.15}
\caption{Detailed comparison experiment results of Agentic category on BFCL-V4. The Agentic category contains two subcategories \textit{i.e.} Web Search and Memory.}
\label{tab:bfcl_agentic}
\resizebox{\linewidth}{!}{
\begin{tabular}{l|c|c|cc|c|ccc}
\toprule
\textbf{Models} & \textbf{Agentic} & \textbf{Web Search} & \textbf{Snippet} & \textbf{No Snippet} & \textbf{Memory} & \textbf{Key Value Store} & \textbf{Vector Store} & \textbf{Rec Sum} \\
\midrule
\multicolumn{9}{l}{{\cellcolor{gray!6}\textit{Open-source Baselines}}}\\
235B-A22B-Instruct & 35.80 & 44.50 & 57.00 & 32.00 & 27.10 & 14.19 & 16.13 & 50.97 \\
30B-A3B-Instruct  & 20.07 & 22.50 & 21.00 & 24.00 & 17.63 & 9.03 & 9.03 & 34.84 \\
4B-Thinking       & 9.19 & 6.00 & 8.00 & 4.00 & 12.37 & 10.65 & 15.81 & 10.65 \\
8B-Thinking       & 13.31 & 12.00 & 15.00 & 9.00 & 14.62 & 5.16 & 7.10 & 31.61 \\
\midrule
\multicolumn{9}{l}{{\cellcolor{gray!6}\textit{From Qwen3-5B-Thinking}}}\\

AUQ                & 11.72 & 3.00 & 4.00 & 2.00 & 20.43 & 13.55 & 19.35 & 28.39 \\
SAGE               & 8.26 & 3.50 & 4.00 & 3.00 & 13.02 & 6.57 & 15.81 & 16.67 \\
Turn-level GRPO    & \underline{13.76} & 3.00 & 4.00 & 2.00 & 24.52 & 12.90 & 20.00 & 40.65 \\
\textbf{Turn-level \sys} & \textbf{13.98} & 3.00 & 4.00 & 2.00 & 24.95 & 12.90 & 23.87 & 38.06 \\ \midrule
\multicolumn{6}{l}{{\cellcolor{gray!6}\textit{From Qwen3-4B-Thinking}}}\\

AUQ                & 14.85 & 14.00 & 17.00 & 11.00 & 15.70 & 9.03 & 14.84 & 23.23 \\
SAGE               & 15.61 & 9.50 & 12.00 & 7.00 & 21.72 & 7.10 & 17.42 & 40.65 \\
\multicolumn{6}{l}{{\cellcolor{gray!6}\textit{From Qwen3-8B-Base}}}\\

Traj.-level CM2    & \underline{21.97} & 22.00 & 31.00 & 13.00 & 21.94 & 9.68 & 14.19 & 41.94 \\
\textbf{Traj.-level \sys} & \textbf{28.79} & 27.00 & 32.00 & 22.00 & 30.57 & 21.00 & 24.55 & 46.16 \\
\bottomrule
\end{tabular}
}
\end{table*}

\begin{table*}[t]
\centering
\caption{Detailed ablation study results on When2Call test set. We do not report results for the base checkpoint of Qwen3-8B-Base since it is not instruction-tuned under this evaluation protocol. The setting w/o $c(s)$ represents $c(s)=1$, \textit{i.e.} $R_{\text{UQ}}=R_{\mathrm{fmt}}+R_{\mathrm{ans}}+R_{\mathrm{cls}}$.}
\label{tab:ablation_results_when2call}
\renewcommand{\arraystretch}{1.15}
\setlength{\tabcolsep}{26pt}
\resizebox{\linewidth}{!}{
\begin{tabular}{l c c c c}
\toprule
\textbf{Models} & \textbf{Acc Norm} & \textbf{F1} & \textbf{Tool Hall} & \textbf{FDAR} \\
\midrule
\multicolumn{5}{l}{{\cellcolor{gray!6}\textit{From Qwen3-4B-Thinking}}}\\
\textbf{Qwen3-4B-Thinking}
& 69.36 & 75.77 & 41.47 & 17.11 \\
\textbf{w/o $R_{\text{ans}}$}
& \underline{78.97} & 79.96 & 38.37 & 2.63 \\
\textbf{w/o $c(s)$}
& 72.46 & 76.06 & 5.73 & 24.76 \\
\textbf{w/o $R_{\text{fmt}}$}
& 75.90 & 76.55 & 24.03 & 1.86 \\
\textbf{Turn-level \sys}
& \textbf{80.83} & 82.84 & 17.83 & 5.07 \\ \hline
\multicolumn{5}{l}{{\cellcolor{gray!6}\textit{From Qwen3-8B-Base}}}\\
\textbf{SFT}
& 19.20 & 27.29 & 16.38 & 63.12 \\
\textbf{+ Traj.-level CM2}
& \underline{43.75} & 36.94 & 75.43 & 9.34 \\
\textbf{+ Traj.-level \sys~ w/o $c(s)$}
& 31.43 & 26.19 & 65.50 & 27.14 \\
\textbf{+ Traj.-level \sys}
& \textbf{62.32} & 60.62 & 49.87 & 8.39 \\
\bottomrule
\end{tabular}
}
\end{table*}

% Table 5
\begin{table*}[t]
\centering
\small
\setlength{\tabcolsep}{16pt}
\renewcommand{\arraystretch}{1.15}
\caption{Detailed ablation study results of trajectory-level post-training on Non-Live category of BFCL-V4.}
\label{tab:bfcl_ablation_non-live}
\resizebox{\linewidth}{!}{
\begin{tabular}{l|c|cccc}
\toprule
\textbf{Models} & \textbf{Non-Live} & \textbf{Simple} & \textbf{Multiple} & \textbf{Parallel} & \textbf{Parallel Multiple} \\
\midrule
SFT              & 79.65 & 69.08 & 89.00 & 80.50 & 80.00 \\
+ CM2              & \underline{80.12} & 70.50 & 93.00 & 85.00 & 72.00 \\
+ \sys~ w/o $c(s)$ & 78.90 & 70.08 & 91.00 & 81.00 & 73.50 \\
+ \sys            & \textbf{82.27} & 71.08 & 94.50 & 85.00 & 78.50 \\
\bottomrule
\end{tabular}
}
\end{table*}

% Table 6
\begin{table*}[t]
\centering
\small
\setlength{\tabcolsep}{16pt}
\renewcommand{\arraystretch}{1.15}
\caption{Detailed ablation study results of trajectory-level post-training on the Live category of BFCL-V4.}
\label{tab:bfcl_ablation_live}
\resizebox{\linewidth}{!}{
\begin{tabular}{l|c|cccc}
\toprule
\textbf{Models} & \textbf{Live} & \textbf{Simple} & \textbf{Multiple} & \textbf{Parallel} & \textbf{Parallel Multiple} \\
\midrule
SFT              & 72.17 & 76.36 & 71.70 & 75.00 & 45.83 \\
+ CM2              & \underline{74.58} & 74.64 & 75.17 & 76.00 & 46.83 \\
+ \sys~ w/o $c(s)$ & 74.17 & 75.58 & 74.36 & 68.75 & 54.17 \\
+ \sys            & \textbf{75.57} & 75.97 & 75.97 & 75.00 & 54.17 \\
\bottomrule
\end{tabular}
}
\end{table*}

% Table 7
\begin{table*}[t]
\centering
\small
\setlength{\tabcolsep}{6pt}
\renewcommand{\arraystretch}{1.15}
\caption{Detailed ablation study results of trajectory-level post-training on the Multi-Turn category of BFCL-V4.}
\label{tab:bfcl_ablation_multi-turn}
\resizebox{\linewidth}{!}{
\begin{tabular}{l|c|cccc}
\toprule
\textbf{Models} & \textbf{Multi-Turn} & \textbf{Multi Turn Base} & \textbf{Multi Turn Missing Function} & \textbf{Multi Turn Missing Parameter} & \textbf{Multi Turn Long Context} \\
\midrule
SFT              & 19.10 & 24.00 & 18.75 & 14.13 & 19.50 \\
+ CM2              & \underline{22.50} & 34.00 & 21.00 & 18.00 & 17.00 \\
+ \sys~ w/o $c(s)$ & 17.37 & 15.50 & 17.50 & 18.50 & 18.00 \\
+ \sys            & \textbf{29.88} & 40.50 & 26.50 & 28.00 & 24.50 \\
\bottomrule
\end{tabular}
}
\end{table*}

% Table 8
\begin{table*}[t]
\centering
\small
\setlength{\tabcolsep}{5pt}
\renewcommand{\arraystretch}{1.15}
\caption{Detailed ablation study results of Agentic category on BFCL-V4.  The Agentic category contains two subcategories \textit{i.e.} Web Search and Memory.}
\label{tab:bfcl_ablation_agentic}
\resizebox{\linewidth}{!}{
\begin{tabular}{l|c|ccccccc}
\toprule
\textbf{Models} & \textbf{Agentic} & \textbf{Web Search} & \textbf{Snippet} & \textbf{No Snippet} & \textbf{Memory} & \textbf{Key Value Store} & \textbf{Vector Store} & \textbf{Rec Sum} \\
\midrule
SFT& 10.87 & 14.00 & 18.00 & 10.00 & 7.74 & 9.03 & 3.87 & 10.32 \\
+ CM2& 21.97 & 22.00 & 31.00 & 13.00 & 21.94 & 9.68 & 14.19 & 41.94 \\
+ \sys~ w/o $c(s)$ & \underline{28.20} & 26.50 & 25.00 & 28.00 & 29.89 & 18.71 & 21.29 & 49.68 \\
+ \sys & \textbf{28.79} & 27.00 & 32.00 & 22.00 & 30.57 & 21.00 & 24.55 & 46.16 \\
\bottomrule
\end{tabular}
}
\end{table*}

% Table 9
\begin{table*}[t]
\centering
\small
\setlength{\tabcolsep}{3pt}
\renewcommand{\arraystretch}{1.15}
\caption{Detailed ablation study results on ToolSandBox.}
\label{tab:toolsandbox_ablation}
\resizebox{\linewidth}{!}{
\begin{tabular}{lcccccccccccccccc}
\toprule
\multirow{2}{*}{\textbf{Model / Method}} & \multirow{2}{*}{\textbf{Overall Score}} & \multicolumn{6}{c}{\textbf{Scenario Categories}} & \multicolumn{7}{c}{\textbf{Tool Augmentations}} \\
\cmidrule(lr){3-8}
\cmidrule(lr){9-17}
 &  & \textbf{STC} & \textbf{MTC} & \textbf{SUT} & \textbf{MUT} & \textbf{SD} & \textbf{C} & \textbf{II} & \textbf{0-DT} & \textbf{3-DT} & \textbf{10-DT} & \textbf{AT} & \textbf{TNS} & \textbf{TDS} & \textbf{ADS} & \textbf{ATS} \\
\midrule
SFT              & 50.42 & 54.02 & 32.73 & 38.78 & 31.42 & 39.07 & 28.70 & 95.70 & 56.10 & 54.88 & 55.46 & 47.39 & 51.61 & 55.46 & 51.57 & 53.34 \\
+ CM2              & \underline{61.21} & 70.57 & 58.55 & 64.22 & 51.93 & 56.64 & 53.70 & 55.60 & 60.12 & 65.04 & 65.02 & 65.03 & 61.80 & 61.30 & 64.14 & 64.51 \\
+ \sys~ w/o $c(s)$ & 56.76 & 54.37 & 40.68 & 44.88 & 39.71 & 43.58 & 38.76 & 96.01 & 61.69 & 61.49 & 62.64 & 61.80 & 61.38 & 61.21 & 61.22 & 61.91 \\
+ \sys  & \textbf{68.28} & 68.66 & 59.70 & 63.10 & 56.93 & 61.58 & 52.04 & 90.70 & 73.01 & 76.41 & 74.94 & 69.16 & 66.74 & 68.77 & 71.26 & 71.24 \\
\bottomrule
\end{tabular}
}
\end{table*}

\end{document}